\pdfoutput=1

\documentclass[11pt]{article}

\usepackage[final]{acl}

\usepackage{times}
\usepackage{latexsym}

\usepackage[T1]{fontenc}

\usepackage[utf8]{inputenc}

\usepackage{microtype}

\usepackage{inconsolata}

\usepackage{graphicx}

\usepackage{mathrsfs}
\usepackage{amsmath}
\usepackage{graphicx}
\usepackage{pgffor}
\usepackage{float}
\usepackage{booktabs}
\usepackage{colortbl}
\usepackage{tabularx}
\usepackage[table]{xcolor}
\usepackage{pbox}
\usepackage{amsfonts, amssymb}
\usepackage{pifont}
\usepackage{cleveref}
\usepackage{hyperref}
\usepackage{makecell}
\usepackage{verbatim}
\usepackage{paralist}
\usepackage[usestackEOL]{stackengine}
\PassOptionsToPackage{table}{xcolor}
\usepackage{mathtools, nccmath}
\usepackage{arydshln}

\usepackage{lipsum}

\usepackage{enumitem}

\usepackage[ruled,vlined]{algorithm2e}
\usepackage{algpseudocode}

\usepackage{xcolor}
\usepackage{xstring}
\usepackage{siunitx}
\usepackage{subcaption}
\usepackage{multirow}
\usepackage{rotating}

\usepackage{nicematrix}

\definecolor{TOKMIX}{HTML}{BBCCEE}
\definecolor{MATHFORM}{HTML}{CCEEFF}
\definecolor{SEQDEP}{HTML}{CCDDAA}
\definecolor{CURRQK}{HTML}{EEEEBB}

\title{Deconstructing Attention: Investigating Design Principles for Effective Language Modeling}

\author{Huiyin Xue, Nafise Sadat Moosavi \and Nikolaos Aletras\\
        School of Computer Science, University of Sheffield, United Kingdom\\
 \texttt{\{hxue12, n.s.moosavi, n.aletras\}@sheffield.ac.uk}}

\begin{document}
\newcommand{\stddev}[1]{%
    \textcolor{gray}{\footnotesize{$_{#1}$}}%
}

\newcommand{\colorednum}[1]{%
    \StrLeft{#1}{1}[\firstchar]%
    \IfStrEq{\firstchar}{+}{%
        \textcolor{teal!80!black!70}{%
        \scriptsize{}}%
    }{%
    \IfStrEq{\firstchar}{-}{%
        \textcolor{red!70!black!70}{%
        \scriptsize{}}%
    }{%
        (#1) 
    }%
    }%
}%

\maketitle
\begin{abstract}

The success of Transformer language models is widely credited to their dot-product attention mechanism, which interweaves a set of key design principles: mixing information across positions (enabling multi-token interactions), sequence-dependent activations (where attention weights adapt to each input), a specific mathematical form (dot-product similarities plus softmax weighting), and coupling of queries and keys to evolving hidden states (grounding attention in the current layer). However, the necessity of each of these principles remains largely untested. In this work, we systematically deconstruct attention by designing controlled variants that selectively relax these principles, applied both uniformly across all layers and in hybrid architectures where only some layers retain standard attention. Our empirical analysis reveals that mechanisms for mixing tokens are indispensable, as their absence collapses models to near-random behavior, while the exact mathematical form and sequence dependency can be substantially relaxed, especially when preserved in just a subset of layers. Surprisingly, even variants that fail in isolation can achieve robust performance when interleaved with standard attention, highlighting a cooperative effect. These findings deepen our understanding of what truly underpins attention’s effectiveness and open new avenues for simplifying language models without sacrificing performance.\footnote{Code is available at \url{https://github.com/HUIYINXUE/DeconAttn}.}

\end{abstract}

\section{Introduction}\label{sec:intro}
The remarkable success of Transformer-based language models~\citep[LMs]{singh2025meta,liu2024deepseek,yang2024qwen2} is widely attributed to the dot-product attention mechanism (i.e. standard attention), which enables these models to weight the significance of each token in a sequence by computing pairwise similarities of their contextual representations~\citep{vaswani2017attention}. However, this powerful mechanism comes at a substantial computational cost with respect to the input sequence length ($L$). This has led to a diverse landscape of proposed mechanisms, including processing longer context~\citep{tay2022efficient}, token-mixing via pooling and multi-layer perceptron MLP-Mixer~\citep{tolstikhin2021mlp}, non-parametric transformations~\citep{yu2022metaformer,lee-thorp-etal-2022-fnet}, optimized kernel functions~\citep{aksenov-etal-2024-linear,arora2024simple,zhen2022cosformer,peng2021random,kasai-etal-2021-finetuning,choromanski2021rethinking,katharopoulos2020transformers}, and linear recurrent neural network (RNN)  architectures~\citep{siems2025deltaproduct,peng2025rwkv,dao2024transformers,yang2024parallelizing,qin2024hgrn,peng2024eagle,poli2023hyena,peng-etal-2023-rwkv,orvieto2023resurrecting}.

Despite this rich body of work, most of these approaches implicitly preserve several underlying design principles inherited from standard attention. Broadly, these principles include: 
(1) incorporating mechanisms for mixing information across tokens (\texttt{Token Mixing}), enabling multi-token interactions, 
(2) emulating the original mathematical form of standard attention (\texttt{Mathematical Form}), i.e. dot-product similarities followed by softmax weighting, 
(3) enforcing strict sequence-dependency in activation maps (\texttt{Sequence-Dependency}), where attention weights depend on the specific input sequence, and 
(4) deriving queries and keys from the current layer's hidden states (\texttt{Current QK}), as opposed to other input types such as uncontextualized representations.
However, the importance of each of these principles remains largely untested. \textit{Are all of these truly essential, or could relaxing some of them suffice if applied selectively?}

Motivated by this foundational question and guided by Occam’s Razor~\citep{sep-simplicity}, we take a diagnostic approach: we systematically relax these principles through controlled attention variants, evaluated in two settings: (1) uniform replacement across all layers, and (2) hybrid configurations that interleave standard and simplified modules.
Through extensive empirical analysis across multiple model sizes, attention variants, and layer configurations, while carefully matching parameter counts of variants, we uncover a set of  insights that refine our understanding of key attention principles. 

Under \textit{uniform} replacement, mechanisms enabling token mixing prove indispensable: removing them, e.g. in \textsl{MLP} variants, leads to near-random accuracy on challenging natural language understanding (NLU) tasks, though such models still capture superficial statistical patterns, as reflected in improved perplexity over trivial baselines. Retaining the dot-product structure and sequence-dependent weighting contributes to stability, but these elements are not strictly necessary in every layer, provided token interactions remain strong.

Notably, in hybrid configurations that interleave simpler attention mechanisms with standard layers, we uncover a striking pattern: attention variants that fail in isolation can nonetheless contribute meaningfully when paired with standard attention, achieving robust performance that often matches or exceeds fully standard models. This suggests standard layers may stabilize activations, mitigate distributional drift, and foster cooperative dynamics across the network, as reflected in both predictive outcomes and structural diagnostics such as attention entropy, head diversity, and sink behaviors.

While hybrid attention schemes have been explored in prior work, such as taking advantages of state space models~\citep{glorioso2024zamba} or augmenting feed-forward modules via mixture-of-experts routing~\citep{lenz2025jamba}, these are typically driven by performance or efficiency goals. \textit{By contrast, our hybrid designs serve as deliberate probes to isolate and examine the causal roles of specific attention properties.} Taken together, our findings challenge the assumption that attention mechanisms must adhere rigidly to their original formulation. By identifying which components are essential and which can be simplified, we outline a path toward new LM architectures that can be structurally leaner and adaptable.

\section{Related Work}
Prior research attributes the success of Transformer models to their efficient token mixing mechanisms. Consequently, numerous studies explore replacing the standard dot-product attention with simpler architectural components that enable parallel training. For instance, \citet{yu2022metaformer} demonstrate the effectiveness of pooling, MLPs, and convolution as alternatives within vision Transformers. Similarly, \citet{lee-thorp-etal-2022-fnet} highlight the efficiency of token mixers based on Fourier transformation and random projection in the BERT model~\citep{devlin-etal-2019-bert}. However, these investigations focus on encoder-only Transformer architectures and may not readily adapt to causal language modeling. While \citet{tolstikhin2021mlp} to introduce a learnable linear layer for token mixing by employing position-wise projection vectors, similar to Linformer~\citep{wang2020linformer}, this approach encounters scalability challenges with long sequences due to its parameter count growing linearly with $L$. Concurrent research largely retains the standard dot-product attention mechanism as a foundational principle. Efforts to reduce the computational cost of this mechanism primarily follow two strategies: weight sharing~\citep{rajabzadeh2024echoatt,ainslie-etal-2023-gqa,xue-aletras-2023-pit,yan2021attention,Kitaev2020Reformer:,shazeer2019fast,xiao2019sharing} or input length shrinkage~\citep{nawrot-etal-2023-efficient,clark-etal-2022-canine,xue-aletras-2022-hashformers}.

Recent work revisits linear RNNs to handle inputs of varying length~\cite{gu2024mamba,poli2023hyena,peng-etal-2023-rwkv,orvieto2023resurrecting,gu2022efficiently}. Follow-up research further improves performance by designing more sophisticated gating mechanisms and update rules~\citep{he2025rodimus,lin2025forgetting,siems2025deltaproduct,peng2025rwkv,dao2024transformers,yang2024parallelizing,qin2024hgrn,peng2024eagle}, with the goal of mimicking human memory, drawing inspiration from the work of \citet{schlag2021linear} on fast weight programmers. Notably, such replacements can also be selectively applied to a subset of attention layers or heads \cite{lenz2025jamba,ren2025vamba,team2024jamba,glorioso2024zamba,peng2024etamba,dong2025hymba,tay-etal-2019-lightweight}. Additionally, this work operates on the contextual representations encoded by deep networks to generate activation maps dynamically.

Another line of research approximates the dot-product computation to achieve linear complexity. These methods rely on various kernel functions that emulate the exponential function using its Taylor expansion~\citep{aksenov-etal-2024-linear,arora2024simple,zhen2022cosformer,peng2021random,kasai-etal-2021-finetuning,choromanski2021rethinking,katharopoulos2020transformers}. This allows for prioritization of the key-value dot product through feature mapping. However, this line of work does not examine the necessity of the other key principles of attention mechanism identified in \S\ref{sec:intro}.

\section{Attention Variants}\label{sec:attn_alternatives}
To operationalize our investigation of the four key design principles identified in \S\ref{sec:intro}, we design targeted variations of attention that selectively relax each property. This allows us to probe their necessity in a controlled, principled framework.

\subsection{Standard Dot-product Attention}

We take standard scaled dot-product attention~\citep{vaswani2017attention} as our baseline, where queries ($\mathbf{Q}$), keys ($\mathbf{K}$), and values ($\mathbf{V}$) are computed from the layer hidden states $\mathbf{H}\in\mathbb{R}^{L\times d_m}$:

{\small
\begin{align}
    \mathbf{O}&=\mathrm{Att}(\mathbf{Q},\mathbf{K},\mathbf{V})=\mathbf{A}\mathbf{V}\\
    \mathbf{A}&=\mathrm{Softmax}\left({\mathbf{Q}\mathbf{K}^\top}/{\sqrt{d_h}}\right)\label{eq:standard_att}
\end{align}
}

{\small
\begin{align}\mathbf{Q},\mathbf{K},\mathbf{V}&=\mathbf{H}\mathbf{W}^{Q,K,V}\label{eq:standard_proj}
\end{align}
}

\noindent This follows all principles: mixing information across positions via $\mathbf{A}$, using a similarity-softmax form, adapting to each input sequence, and tying $\mathbf{Q}$, $\mathbf{K}$ to the current hidden state $\mathbf{H}$.

\subsection{Relaxing Token Mixing}
\paragraph{MLP.}
To directly examine the necessity of cross-token interactions, we replace attention with a gated MLP layer, consisting of three fully-connected (FC) layers ($\mathrm{FC_{Dn}}, \mathrm{FC_{Gt}}, \mathrm{FC_{Up}}$) for down-projection, gating and down-projection respectively. This  effectively eliminates token mixing and each token is processed independently, only attending to itself.

{\small
\begin{align}
\mathbf{O}&=\mathrm{GatedMLP}(\mathbf{H})\\
&=\mathrm{FC_{Dn}}(\mathrm{SiLU}(\mathrm{FC_{Gt}}(\mathbf{H}))\cdot\mathrm{FC_{Up}}(\mathbf{H}))\label{eq:gated_mlp}
\end{align}
}

\noindent We use a SiLU activation~\citep{elfwing2018sigmoid} and match the parameter count of standard attention. This variant serves as a minimal baseline to assess how much attention's effectiveness depends on cross-token interaction, beyond what feed-forward paths alone can provide without using any $\mathbf{Q}$, $\mathbf{K}$ and $\mathbf{V}$.

\subsection{Relaxing the Mathematical Form}
We assess whether attention must strictly follow the canonical dot-product plus softmax formulation. To this end, we evaluate two variants that either approximate or break this form.

\paragraph{Approximate.}
Following \citet{arora2024simple}, we preserve the mathematical intention of similarity-based weighting, while relaxing the exact form of softmax via a second-order Taylor expansion, yielding a linear-time recurrent form (Appx.~\ref{appendix:recurrent_form}):

{\small
\begin{align}
\mathbf{A}&\approx\mathrm{Taylor}\left({\mathbf{Q}\mathbf{K}^\top}/{\sqrt{d_h}}\right)\label{eq:approx_attn}
\end{align}
}

\noindent $\mathbf{Q}$, $\mathbf{K}$ and $\mathbf{V}$ are computed using Eq.~\ref{eq:standard_proj}.

\paragraph{Non-approximate.}
To contrast this, we introduce a new variant that discards explicit pairwise similarity altogether. Instead of computing an attention matrix via $\mathbf{QK}^\top$, it uses element-wise self-gating, multiplying $\mathbf{Q}$ and $\mathbf{K}$ derived from the same hidden state, and normalizes the result across time steps with softmax:

{\small
\begin{align}
\mathbf{A}
&=\mathrm{Softmax}\left({(\mathbf{Q}\odot\mathbf{K})\mathbf{1}^\top}/{\sqrt{d_h}}\right)\label{eq:non_approx_attn}\\
\mathbf{Q}&=\mathrm{SiLU}\left(\mathbf{H}\mathbf{W}^Q\right);\quad\mathbf{K},\mathbf{V}=\mathbf{H}\mathbf{W}^{K,V}\label{eq:non_approx_x}
\end{align}
}

\noindent This variant follows an entirely different mathematical form to standard attention. We expect that this should make it harder for adjacent context tokens to receive large attention scores, as the denominator in the softmax computation monotonically increases (see recurrent form in Appx.~\ref{appendix:recurrent_form}). Notably, the SiLU activation is applied element-wise and does not introduce additional complexity. We apply SiLU activation on $\mathbf{Q}$ projection to add non-linearity. This does not require additional parameters and allows parallelism during training.

\subsection{Relaxing Sequence Dependency}\label{sec:modified_qk}

To test whether attention weights must be dynamically adapted to each input sequence (i.e. sequence-dependent), we construct two variants where $\mathbf{Q}$ and $\mathbf{K}$ are fixed across all inputs, inspired by MLP-Mixer~\citep{fusco-etal-2023-pnlp,tolstikhin2021mlp}, but making the parameter count in attention blocks independent of the maximum sequence length. Relaxing sequence dependency allows attention scores for all inputs to be pre-computed and cached during inference.

\paragraph{Random-fixed (RndEmbQK).}

We initialize a set of random embeddings $\boldsymbol{\epsilon} \sim \mathcal{N}(0, \sigma^2 \mathbf{I})$ that remain constant across inputs. These are passed through the Transformer stack up to layer $l$:

{\small
\begin{align}
\mathbf{X} &= \mathrm{TransformerBlock}^{(l)}(\mathbf{\mathbf{\epsilon}}), \quad
\mathbf{\mathbf{\epsilon}}\sim\mathcal{N}(\mathbf{0},\sigma\mathbf{I})\label{eq:random_1}
\end{align}
}

{\small
\begin{align}\mathbf{Q},\mathbf{K} = \mathbf{X}\mathbf{W}^{Q,K};\quad \mathbf{V} = \mathbf{H}\mathbf{W}^V\label{eq:random_2}
\end{align}
}

\noindent Since $\mathbf{Q}$ and $\mathbf{K}$ do not depend on the input, attention maps are fixed and do not adapt to context.

\paragraph{Text-fixed (FixedSeqQK).}

Instead of random embeddings, we use a randomly selected fixed sequence of natural language tokens $\mathbf{t}^s$ (first 2048 tokens from FineWeb-10BT~\citep{lozhkov2024fineweb-edu}). These are embedded and passed through the Transformer to generate $\mathbf{X}$:

{\small
\begin{align}
\mathbf{X} &= \mathrm{TransformerBlock}^{(l)}(\mathrm{Emb}(\mathbf{t}^s))
\end{align}
}

{\small
\begin{align}\mathbf{Q},\mathbf{K} = \mathbf{X}\mathbf{W}^{Q,K};\quad \mathbf{V} = \mathbf{H}\mathbf{W}^V\label{eq:pseudo_2}
\end{align}
}

\noindent This setup also produces fixed attention maps, but grounded in natural text instead of completely randomly initialized embeddings. Compared to \textsl{RndEmbQK}, it may encode weak structural priors, such as grammatical patterns or token co-occurrences.
These variants allow us to test whether dynamic, input-conditioned attention maps are necessary, or whether fixed maps, paired with learned value paths, are sufficient.

\subsection{Relaxing the Derivation of Q and K}
\paragraph{StaticEmbQK.}
Finally, to test whether tying $\mathbf{Q},\mathbf{K}$ to current layer hidden states ($\mathbf{H}$ or $\mathbf{X}$ above) is essential, we compute them directly from static input embeddings $\mathbf{e}$ corresponding to the input sequence $\mathbf{t}$:

{\small
\begin{align}
\mathbf{Q},\mathbf{K} = \mathbf{e}\mathbf{W}^{Q,K};\quad \mathbf{V} = \mathbf{H}\mathbf{W}^V;\quad \mathbf{e}=\mathrm{Emb}(\mathbf{t})
\end{align}
}

\noindent This means that while attention maps are not fixed, they are computed without contextualization from the evolving hidden representations. It further allows attention scores from different layers to be computed in parallel.

\section{Experimental Setup}
\subsection{Data}
We use seven zero-shot NLU tasks in English:  \textsc{ARC-E}~\citep{clark2018think}, \textsc{BoolQ}~\citep{clark-etal-2019-boolq}, \textsc{COPA}~\citep{roemmele2011choice}, \textsc{PiQA}~\citep{bisk2020piqa}, \textsc{SciQ}~\citep{welbl-etal-2017-crowdsourcing}, \textsc{RTE}~\citep{wang2019superglue} and \textsc{HellaSwag}~\citep{zellers-etal-2019-hellaswag}. We also experiment with two LM tasks: \textsc{WikiText}~\citep{merity2017pointer} and \textsc{Lambada Openai}~\citep{radford2019language}.

\subsection{Implementation Details}\label{sec:implementation_details}

\paragraph{Base model.} Our models are built upon Qwen2.5~\citep{yang2024qwen2}. However, we replace its standard attention mechanism with the alternative attention modules detailed in \S~\ref{sec:attn_alternatives}. To ensure a strict parameter count match across all attention variants, we use multi-head attention \citep{vaswani2017attention}, deviating from Qwen2.5's default grouped-query attention~\citep{ainslie-etal-2023-gqa}. For tokenization, we use the 50K English-centric BPE~\citep{sennrich-etal-2016-neural} vocabulary of Pythia~\citep{biderman2023pythia}, offering small memory footprint, and fast training.

\paragraph{Model configurations.} We pretrain  models with approximately 500M parameters, using two  configurations: (1) \textit{Uniform} with simple attention mechanisms across all Transformer layers; (2) \textit{Hybrid} that integrates simple attention mechanisms in odd-numbered layers and standard attention in even-numbered layers. To assess the contribution of the modified attention variants within the \textit{hybrid} configuration, we introduce a configuration where we remove the odd-numbered layers from pre-trained \textit{hybrid} models (\textit{skip}) and evaluate the resulting performance without additional training. 

We further test these three configurations by training models of 70 million and 160 million parameters (see Appx.~\ref{appendix:scaling_trend}). We finally explore various alternative \textit{hybrid} configurations such as changing the simple attention replacement ratio, the details of which are presented in \S~\ref{sec:different_hybrid_config}. Specific model size details are provided in Appx.~\ref{appendix:model_config_appendix}. Meanwhile, we strictly constrain all models with different attention variants to have the same number of parameters to eliminate any effects from differences in size.

\paragraph{Pre-training.} All models are pre-trained on the SlimPajama dataset~\citep{cerebras2023slimpajama} for up to 15 billion tokens, following Chinchilla scaling laws~\citep{hoffmann2022training}. We use a mini-batch size of 500K tokens, aligning with the training budget outlined in Titans~\citep{behrouz2024titans}. To optimize pre-training efficiency, we use a sequence length of 2048 tokens.\footnote{Details on hyperparameter selection is provided in Appx.~\ref{appendix:hyperparameters}. For both pre-training and evaluation, we use a single AMD Instinct MI300X accelerator.}

\subsection{Predictive Performance Evaluation}
We use the LM-evaluation-harness toolkit v0.4.8~\citep{eval-harness} for evaluation. We report accuracy for all NLU tasks and perplexity (PPL) for LM tasks. For \textsc{Lambada Openai}, we report both.

\subsection{Attention Pattern Indicators}

Looking at the performance itself may not offer a comprehensive picture of the behavior of the different attention mechanisms we test. To obtain a more granular understanding of their internal workings, we investigate their attention patterns. We compute eight indicators from the attention matrices $\mathbf{A}_j \in \mathbb{R}^{L \times L}$ for each head $j=1, \ldots, n_h$ in a given layer. We specifically focus on \textit{attention sinks}, i.e. over-attending to the initial token in a sequence, and \textit{local patterns} within attention matrices, i.e. prioritizing nearby tokens, following prior work~\citep{xiao2024efficient, han-etal-2024-lm}.\footnote{\textsc{Entropy (H)}, \textsc{Conc}, and \textsc{HeadDiv} are min-max normalized. \textsc{Sink} and \textsc{LocFoc$N$} use absolute values (\textsc{LocFoc$N$} is scaled by two for visibility). High \textsc{Entropy} and low \textsc{Conc} suggest mean-pooling like behavior. High \textsc{Conc} and low \textsc{Entropy} indicate focus on a few tokens. Further examination of \textsc{Sink} and \textsc{LocFoc$N$} clarifies if this focus is on the first token or local tokens. Low \textsc{Entropy} and high \textsc{Conc} with low scores elsewhere (except \textsc{HeadDiv}) may point to sparse attention on mid-sequence tokens.}

\paragraph{Entropy (H).} Measures the randomness of attention scores. Higher \textsc{Entropy} indicates more uniform attention distribution across tokens, similar to mean-pooling: $\mathrm{H} = -\sum_{a \in \mathbf{A}} a \cdot \log(a)$.

\paragraph{Concentration (Conc).} Measures the concentration of attention. A higher Frobenius norm $\lVert\mathbf{A}\rVert_F$ indicates attention is focused on a limited number of tokens: $ \mathrm{Conc} = \lVert\mathbf{A}\rVert_F = \sqrt{\sum_{a \in \mathbf{A}} a^2}$.

\paragraph{Head diversity (HeadDiv).} Quantifies the variability of attention patterns across different heads. Calculated as the average position-wise standard deviation across heads, higher \textsc{HeadDiv} suggests better use of the multi-head mechanism.

{\small
$$ \mathrm{HeadDiv} = \dfrac{2}{{L(1+L)}}{\sum \mathrm{std}(\{\mathbf{A}_1, \ldots, \mathbf{A}_{n_h}\})} $$
}

\paragraph{Attention sink (Sink).} Detects focus on the first token. It is the average attention score assigned by all queries to the initial token. Higher Sink means a stronger attention sink: $\mathrm{Sink} = {\sum \mathbf{A}_{:,1}}/{L}$.

\paragraph{Local Focus (LocFocN).} Measures the attention focus on nearby tokens. It is the average attention score for tokens at a fixed relative distance $N$ (here $N \in \{0, 1, 2, 3\}$). Higher LocFocN suggests stronger contribution from local context.

{\small
$$ \mathrm{LocFocN} = {\sum \mathbf{A}_{L-N, L-N}}/\left({L-N}\right) $$
}



\begin{table*}[!t]
\centering
\small
\resizebox{1.0\linewidth}{!}{
\begin{NiceTabular}{clccccccccrrr}
\toprule
& & \textbf{ARC-E} & \textbf{BoolQ} & \textbf{COPA} & \textbf{PiQA} & \textbf{SciQ} & \textbf{RTE} & \textbf{HellaSwag} & \text{Avg.} & \textbf{Wiki} & \multicolumn{2}{c}{\textbf{LAMBADA}}\\ 
& & acc$\uparrow$ & acc$\uparrow$ & acc$\uparrow$ & acc$\uparrow$ & acc$\uparrow$ & acc$\uparrow$ & acc$\uparrow$ & acc$\uparrow$ & ppl$\downarrow$ & ppl$\downarrow$ & acc$\uparrow$ \\ 
 \cmidrule(lr){3-10}\cmidrule(lr){11-13}

&\textbf{Rnd. Guess} & 25.0\stddev{0.0} & 50.0\stddev{0.0} & 50.0\stddev{0.0} & 50.0\stddev{0.0} & 25.0\stddev{0.0} & 50.0\stddev{0.0} & 25.0\stddev{0.0} & 39.9 & 3E+5 & 3E+6 & 0.0\stddev{0.0} \\ 
&\textbf{Majority} & 25.7\stddev{0.0} & 62.2\stddev{0.0} & 56.0\stddev{0.0} & 50.5\stddev{0.0} & 25.0\stddev{0.0} & 52.7\stddev{0.0} & 25.0\stddev{0.0} & 39.9 & - & - & -\\ \hdashline

&Standard & 41.5\stddev{1.0} & 56.6\stddev{0.9} & 63.0\stddev{4.9} & 60.9\stddev{1.1} & 60.2\stddev{1.5} & 53.1\stddev{3.0} & 28.3\stddev{0.4} & 51.9 & \textbf{38.1} & 134.1 & 22.9\stddev{0.5
}\\ \midrule

\rowcolor{TOKMIX} \cellcolor{nocolor}{\multirow{6}{*}{\rotatebox[origin=c]{90}{UNIFORM}}}
&MLP & 28.5\stddev{0.9}\colorednum{-13.0} & 37.8\stddev{0.8}\colorednum{-18.8} & 54.0\stddev{5.0}\colorednum{- 9.0} & 54.8\stddev{1.2}\colorednum{-6.1} & 25.9\stddev{1.4}\colorednum{-34.3} & 52.7\stddev{3.0}\colorednum{-0.4} & 26.1\stddev{0.4}\colorednum{-2.2} & 40.0\colorednum{-23.0} & 993.5 & 1E+5 & 0.0\stddev{0.0}\\
\rowcolor{MATHFORM} \cellcolor{nocolor}&Approx. & 40.7\stddev{1.0}\colorednum{- 0.8} & 51.5\stddev{0.9}\colorednum{- 5.1} & 64.0\stddev{4.8}\colorednum{+ 1.0} & 59.9\stddev{1.1}\colorednum{-1.0} & 55.0\stddev{1.6}\colorednum{- 5.2} & 52.3\stddev{3.0}\colorednum{-0.8} & 28.1\stddev{0.4}\colorednum{-0.2} & 50.2\colorednum{- 3.3} & 47.9 & 238.6 & 18.5\stddev{0.5}\\
\rowcolor{MATHFORM} \cellcolor{nocolor}&Non-apx. & 26.8\stddev{0.9}\colorednum{-14.7} & 37.8\stddev{0.8}\colorednum{-18.8} & 60.0\stddev{4.9}\colorednum{- 3.0} & 53.2\stddev{1.2}\colorednum{-7.7} & 19.3\stddev{1.2}\colorednum{-40.9} & 52.3\stddev{3.0}\colorednum{-0.8} & 26.0\stddev{0.4}\colorednum{-2.3} & 39.3\colorednum{-24.3} & 9E+4 & 2E+6 & 0.0\stddev{0.0}\\
\rowcolor{SEQDEP} \cellcolor{nocolor}&RndEmbQK & 39.5\stddev{1.0}\colorednum{- 2.0} & 55.3\stddev{0.9}\colorednum{- 1.3} & 57.0\stddev{5.0}\colorednum{- 6.0} & 59.8\stddev{1.1}\colorednum{-1.1} & 46.4\stddev{1.6}\colorednum{-13.8} & 50.9\stddev{3.0}\colorednum{-2.2} & 27.2\stddev{0.4}\colorednum{-1.1} & 48.0\colorednum{- 7.6} & 84.8 & 6402.4 & 1.3\stddev{0.2}\\
\rowcolor{SEQDEP} \cellcolor{nocolor}&FixedSeqQK & 39.4\stddev{1.0}\colorednum{- 2.1} & \textbf{59.0}\stddev{0.9}\colorednum{+ 2.4} & 61.0\stddev{4.9}\colorednum{- 2.0} & 59.4\stddev{1.1}\colorednum{-1.5} & 51.2\stddev{1.6}\colorednum{- 9.0} & 52.7\stddev{3.0}\colorednum{-0.4} & 27.5\stddev{0.4}\colorednum{-0.8} & 50.0\colorednum{- 3.7} & 79.1 & 19578.1 & 1.4\stddev{0.2}\\
\rowcolor{CURRQK} \cellcolor{nocolor}&StaticEmbQK & 39.6\stddev{1.0}\colorednum{- 1.9} & 52.9\stddev{0.9}\colorednum{- 3.7} & 63.0\stddev{4.9}\colorednum{+ 0.0} & 59.4\stddev{1.1}\colorednum{-1.5} & 49.2\stddev{1.6}\colorednum{-11.0} & 54.2\stddev{3.0}\colorednum{+1.1} & 27.2\stddev{0.4}\colorednum{-1.1} & 49.4\colorednum{- 5.0} & 79.9 & 2287.4 & 3.3\stddev{0.2}\\
\midrule

\rowcolor{TOKMIX} \cellcolor{nocolor}\multirow{6}{*}{\rotatebox[origin=c]{90}{HYBRID}}
&MLP & 37.5\stddev{1.0}\colorednum{- 4.0} & 49.8\stddev{0.9}\colorednum{- 6.8} & 60.0\stddev{4.9}\colorednum{- 3.0} & 60.2\stddev{1.1}\colorednum{-0.7} & 54.3\stddev{1.6}\colorednum{- 5.9} & 52.7\stddev{3.0}\colorednum{-0.4} & 26.1\stddev{0.4}\colorednum{-2.2} & 48.7\colorednum{- 6.3} & 45.8 & 228.7 & 20.8\stddev{0.6}\\
\rowcolor{MATHFORM} \cellcolor{nocolor}&Approx. & 39.9\stddev{1.0}\colorednum{- 1.6} & 51.5\stddev{0.9}\colorednum{- 5.1} & \textbf{67.0}\stddev{4.7}\colorednum{+ 4.0} & 60.4\stddev{1.1}\colorednum{-0.5} & 60.5\stddev{1.5}\colorednum{+0.3} & 53.4\stddev{3.0}\colorednum{+ 0.3} & 28.4\stddev{0.4}\colorednum{+0.1} & 51.6\colorednum{- 0.7} & 39.4 & 140.0 & 23.7\stddev{0.6}\\
\rowcolor{MATHFORM} \cellcolor{nocolor}&Non-apx. & \textbf{42.3}\stddev{1.0}\colorednum{+ 0.8} & 56.8\stddev{0.9}\colorednum{+ 0.2} & 63.0\stddev{4.9}\colorednum{+ 0.0} & 61.7\stddev{1.1}\colorednum{+0.8} & \textbf{63.0}\stddev{1.5}\colorednum{+ 2.8} & 54.9\stddev{3.0}\colorednum{+1.8} & \textbf{28.5}\stddev{0.5}\colorednum{+0.2} & \textbf{52.9}\colorednum{+1.8} & 39.4 & \textbf{133.1} & \textbf{23.8}\stddev{0.6}\\
\rowcolor{SEQDEP} \cellcolor{nocolor}&RndEmbQK & 40.1\stddev{1.0}\colorednum{- 1.4} & 48.3\stddev{0.9}\colorednum{- 8.3} & 61.0\stddev{4.9}\colorednum{- 2.0} & 61.2\stddev{1.1}\colorednum{+0.3} & 60.0\stddev{1.5}\colorednum{- 0.2} & 50.9\stddev{3.0}\colorednum{-2.2} & 27.2\stddev{0.4}\colorednum{-1.1} & 49.8\colorednum{- 4.1} & 39.3 & 157.5 & 22.0\stddev{0.6}\\
\rowcolor{SEQDEP} \cellcolor{nocolor}&FixedSeqQK & 40.5\stddev{1.0}\colorednum{- 1.0} & 58.5\stddev{0.9}\colorednum{+ 1.9} & 64.0\stddev{4.8}\colorednum{+ 1.0} & \textbf{61.9}\stddev{1.1}\colorednum{+1.0} & 62.0\stddev{1.5}\colorednum{+ 1.8} & 52.7\stddev{3.0}\colorednum{-0.4} & 28.4\stddev{0.4}\colorednum{+0.1} & 52.6\colorednum{+ 1.2} & 38.5 & 354.7 & 20.3\stddev{0.6}\\
\rowcolor{CURRQK} \cellcolor{nocolor}&StaticEmbQK & 39.2\stddev{1.0}\colorednum{- 2.3} & 54.7\stddev{0.9}\colorednum{- 1.9} & 64.0\stddev{4.8}\colorednum{+ 1.0} & 60.9\stddev{1.1}\colorednum{+0.0} & 58.4\stddev{1.6}\colorednum{- 1.8} & \textbf{57.4}\stddev{3.0}\colorednum{+4.3} & 28.2\stddev{0.4}\colorednum{-0.1} & 51.8\colorednum{- 0.2} & 38.7 & 140.7 & \textbf{23.8}\stddev{0.6}\\

\midrule

\multirow{6}{*}{\rotatebox[origin=c]{90}{HYBRID SKIP}}
\rowcolor{TOKMIX} \cellcolor{nocolor}&MLP & 24.4\stddev{0.9}\colorednum{-17.1} & 41.8\stddev{0.9}\colorednum{-14.8} & 54.0\stddev{5.0}\colorednum{- 9.0} & 52.8\stddev{1.2}\colorednum{-8.1} & 19.0\stddev{1.2}\colorednum{-41.2} & 46.9\stddev{1.7}\colorednum{-6.2} & 25.6\stddev{0.4}\colorednum{-2.7} & 37.8\colorednum{-27.3} & 2E+5 & 5E+6 & 0.0\stddev{0.0}\\
\rowcolor{MATHFORM} \cellcolor{nocolor}&Approx. & 26.6\stddev{0.9}\colorednum{-14.9} & 46.1\stddev{0.9}\colorednum{-10.5} & 59.0\stddev{4.9}\colorednum{- 4.0} & 52.8\stddev{1.2}\colorednum{-8.1} & 20.1\stddev{1.3}\colorednum{-40.1} & 48.0\stddev{3.0}\colorednum{-5.1} & 26.0\stddev{0.4}\colorednum{-2.3} & 39.8\colorednum{-23.4} & 2E+6 & 1E+7 & 0.0\stddev{0.0}\\
\rowcolor{MATHFORM} \cellcolor{nocolor}&Non-apx. & 26.6\stddev{0.9}\colorednum{-14.9} & 39.2\stddev{0.9}\colorednum{-17.4} & 52.0\stddev{5.0}\colorednum{-11.0} & 51.4\stddev{1.2}\colorednum{-9.5} & 20.4\stddev{1.3}\colorednum{-39.8} & 46.9\stddev{3.0}\colorednum{-6.2} & 25.8\stddev{0.4}\colorednum{-2.5} & 37.5\colorednum{-27.9} & 5E+5 & 9E+6 & 0.0\stddev{0.0}\\
\rowcolor{SEQDEP} \cellcolor{nocolor}&RndEmbQK & 27.4\stddev{0.9}\colorednum{-14.1} & 37.8\stddev{0.8}\colorednum{-18.8} & 58.0\stddev{5.0}\colorednum{- 5.0} & 53.3\stddev{1.2}\colorednum{-7.6} & 21.1\stddev{1.3}\colorednum{-39.1} & 52.7\stddev{3.0}\colorednum{-0.4} & 26.1\stddev{0.4}\colorednum{-2.2} & 39.5\colorednum{-24.0} & 2E+4 & 3E+6 & 0.0\stddev{0.0}\\
\rowcolor{SEQDEP} \cellcolor{nocolor}&FixedSeqQK & 27.2\stddev{0.9}\colorednum{-14.3} & 39.4\stddev{0.9}\colorednum{-17.2} & 59.0\stddev{4.9}\colorednum{- 4.0} & 52.3\stddev{1.2}\colorednum{-8.6} & 22.1\stddev{1.3}\colorednum{-38.1} & 48.4\stddev{3.0}\colorednum{-4.7} & 25.9\stddev{0.4}\colorednum{-2.4} & 39.2\colorednum{-24.6\%} & 2E+5 & 5E+6 & 0.0\stddev{0.0}\\
\rowcolor{CURRQK} \cellcolor{nocolor}&StaticEmbQK & 25.5\stddev{0.9}\colorednum{-16.0} & 43.0\stddev{0.9}\colorednum{-13.6} & 57.0\stddev{5.0}\colorednum{- 6.0} & 53.1\stddev{1.2}\colorednum{-7.8} & 22.0\stddev{1.3}\colorednum{-38.2} & 51.6\stddev{3.0}\colorednum{-1.5} & 25.9\stddev{0.4}\colorednum{-2.4} & 39.7\colorednum{-23.5\%} & 7E+4 & 5E+6 & 0.0\stddev{0.0}\\

\bottomrule

\end{NiceTabular}
}
\caption{Performance of \textit{uniform}, \textit{hybrid}, \textit{skip} and \textsl{standard} models (500M). Purple (MLP), blue (Approx., Non-apx.), green (RndEmbQK, FixedSeqQK) and yellow (StaticEmbQK) denote variants that relax \texttt{Token Mixing}, \texttt{Mathematical Form}, \texttt{Sequence-Dependency} and  \texttt{Current QK}, respectively.}
\label{table:down_stream_performance}

\end{table*}

\section{Results}

Tbl.~\ref{table:down_stream_performance} shows the performance of all model variants (\S\ref{sec:attn_alternatives}), employing \textit{uniform}, \textit{hybrid}, and \textit{skip} configurations across NLU and LM tasks. Results illuminate the role each design principle plays in effective language modeling.

\paragraph{Token mixing is crucial.} 
The uniform MLP model, which lacks any cross-token interaction, performs near chance on most NLU tasks, highlighting that token mixing is essential for reasoning and understanding. Despite this, it achieves a much lower perplexity on WikiText (993.5 vs. ~300K for \textsl{RndEmbQK}), indicating that even without explicit mixing, MLP can memorize or exploit local token statistics, likely unigram or bigram patterns.
Introducing token mixing in a hybrid setup substantially improves NLU performance (e.g. 9.2 average accuracy points over uniform MLP), showing that mixing in part of the network can compensate to a degree. Still, the hybrid MLP variant has the highest WikiText perplexity among all hybrids, indicating that token mixing across all layers is important for fully modeling long-range dependencies.

\paragraph{Standard mathematical form is important in uniform.} When applied uniformly, variants that retain the core structure of attention (e.g. \textsl{Approximate}, \textsl{RndEmbQK}, \textsl{FixedSeqQK} and \textsl{StaticEmbQK}) restore over 92\% of the average NLU accuracy of attention. In contrast, \textsl{Non-approximate}, which discards this structure, performs close to random guess (39.3 vs. 39.9 on NLU Avg. accuracy). \textsl{Approximate} achieves the strongest results among uniform variants (8.8 higher PPL on WikiText), suggesting that preserving or closely approximating its mathematical form appears critical for maintaining predictive performance.

\paragraph{Sequence-dependency enhances the generalization ability.} 
To assess the role of sequence-dependent attention, we compare variants that retain similar architectures but differ in whether attention scores vary across inputs. \textsl{StaticEmbQK}, which preserves \texttt{Sequence-Dependency}, consistently outperforms \textsl{RndEmbQK} and \textsl{FixedSeqQK}, which use fixed attention patterns, particularly on \textsc{Lambada Openai} by around 2\% higher accuracy.
This pattern holds across both uniform and hybrid settings. Additionally, hybrid models that preserve sequence-dependency, such as \textsl{Approximate}, \textsl{StaticEmbQK}, and \textsl{Non-approximate}, tend to perform better on global-context benchmarks. These results suggest that input-specific attention contributes to better generalization, even when other attention properties are simplified.

\paragraph{Current QK is not as essential as expected.} \textsl{StaticEmbQK} relaxes \texttt{Current QK}. Though it does not match the PPL of \textsl{standard} across language modeling tasks, it results in PPL of 79.9 twice as high as 38.1 of \textsl{standard} under \textit{uniform} configuration on \textsc{WikiText}. It also  greatly outperforms \textsl{MLP}, reducing PPL tenfold (from 993.5 on \textsc{WikiText}), while its predictive performance is comparable to \textsl{standard}. Moreover, under \textit{hybrid} configuration, it achieves predictive performance comparable to \textsl{standard} baseline across all tasks. It indicates \texttt{Current QK} is not as essential for strong predictive performance as initially believed.

\paragraph{Layer collaboration matters.} All \textit{hybrid} models where simple attention variants are used in odd layers and standard attention in even layers achieve predictive performance comparable to \textsl{Standard} attention on both NLU and language modeling tasks. Surprisingly, \textsl{Non-approximate} attention, the worst performer in the uniform configuration, demonstrates strong performance in this \textit{hybrid} setup, slightly surpassing \textsl{Standard} on average NLU accuracy (+1.8\%) and \textsc{Lambada Openai} accuracy (+0.9\%), while reducing PPL by 1.0. The \textit{hybrid} configuration also alleviates the relatively higher uncertainty observed with \textsl{RndEmbQK} and \textsl{FixedSeqQK}, halving their \textsc{WikiText} PPL by incorporating standard layers that aid in grounding attention to individual inputs. These findings suggest that layers exhibiting poor performance in isolation can be effective when combined with stronger layers (i.e. standard attention).

Considering the residual connections, which facilitate information flow along a shortcut pathway bypassing the simple attention alternatives, we further conduct an ablation study to constrain information flow solely through these residual connections. This involves skipping the non-\textsl{Standard} layers when pre-training \texttt{hybrid} models (denoted as \texttt{SKIP} in Tbl.~\ref{table:down_stream_performance}). The results provide further support to the assumption of layer collaboration. All variants in \texttt{w/ SKIP} perform even slightly worse than random guessing (i.e. average accuracy lower than 39.9 on NLU) and further result in PPL explosion in language modeling compared to \texttt{hybrid} by a margin. This indicates that the non-\textsl{Standard} layers, despite their simplicity or poor performance in uniform configurations, contribute positively to the overall predictive performance in hybrid architectures.

\begin{figure}[!t]
\centering
    \begin{subfigure}[t]{0.48\textwidth}
        \centering
        \includegraphics[width=1.0\textwidth]{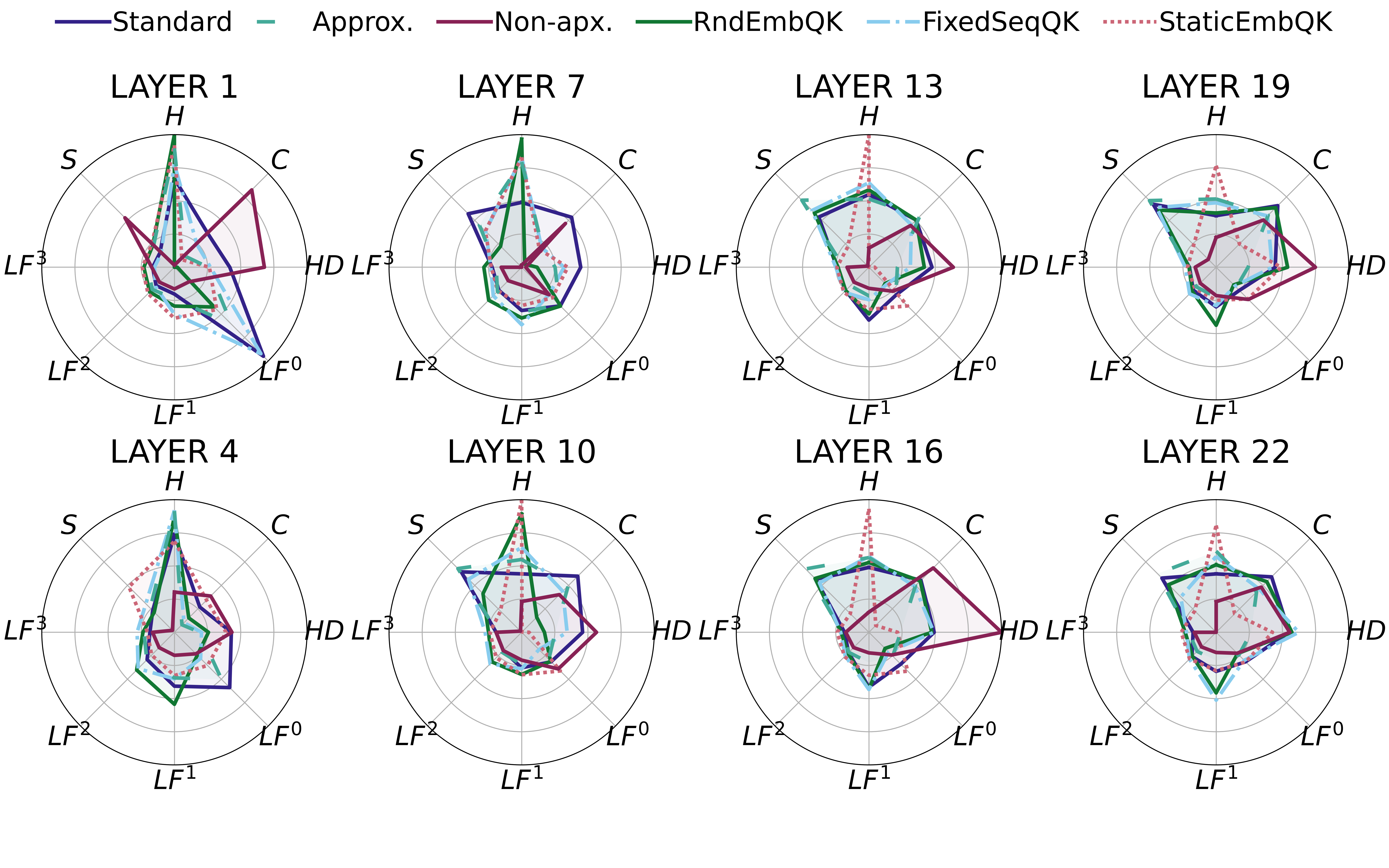}
    \end{subfigure}%
    \\
    \begin{subfigure}[t]{0.48\textwidth}
        \centering
        \includegraphics[width=1.0\textwidth]{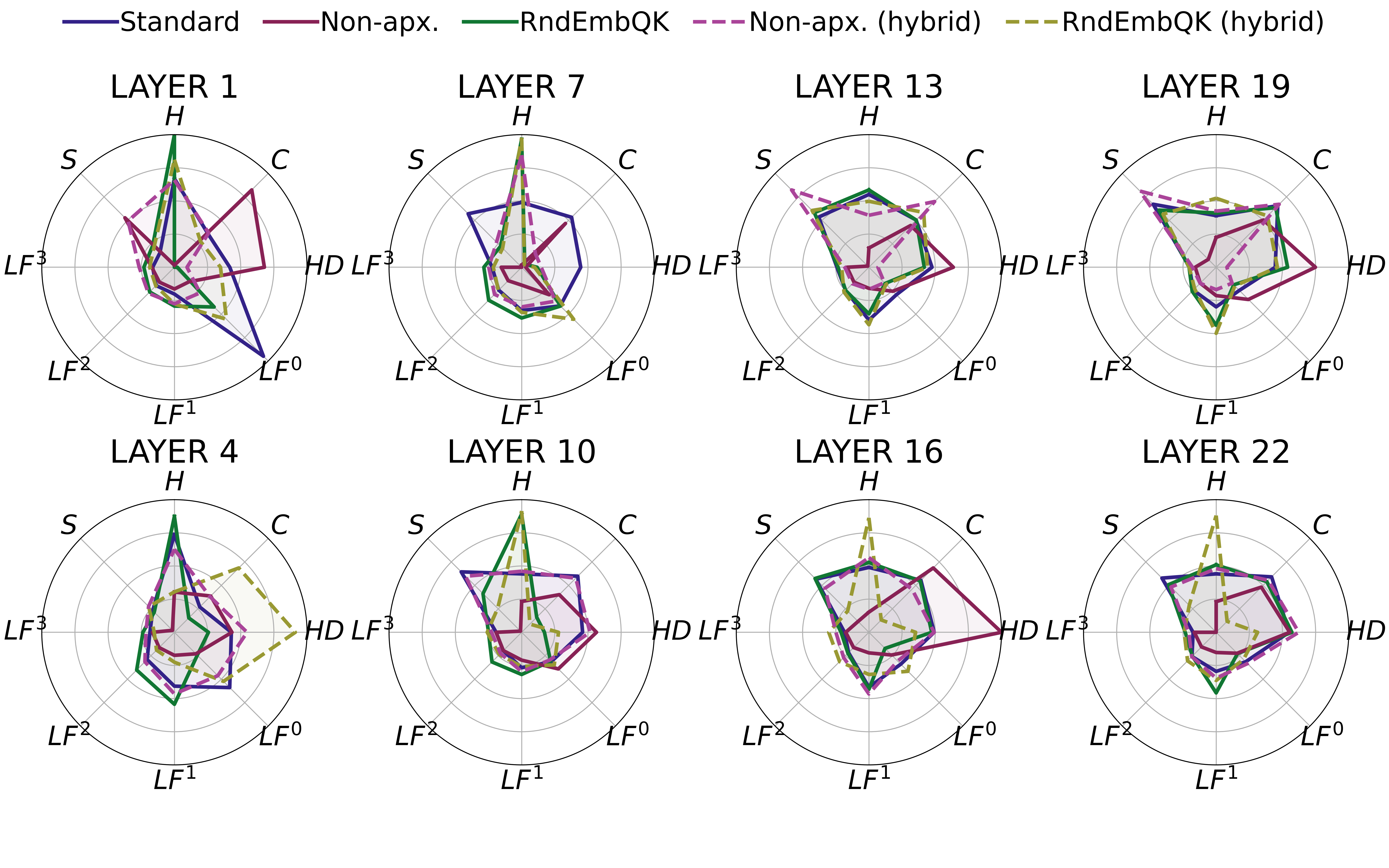}
    \end{subfigure}
\caption{Layer-wise attention indicators for \textsl{Approx.}, \textsl{Non-approx.}, \textsl{RndEmbQK}, \textsl{FixedSeqQK} and \textsl{StaticEmbQK} in \textit{uniform} (top) and \textit{hybrid} (bottom) configurations, and \textsl{Standard} ($H$: \textsc{Entropy}, $C$: \textsc{Conc}, $HD$: \textsc{HeadDiv}, $LF$: \textsc{LocFoc$N$}, $S$: \textsc{Sink}).}
\label{fig:radar_24}
\end{figure}

\section{Analysis and Discussion}
\paragraph{Attention variants.}

\textsl{Non-approximate} attention that relaxes standard attention's \texttt{Mathematical Form} appears to be the most challenging to train in a \textit{uniform} configuration. Radar plots in Fig.~\ref{fig:radar_24} show very low \textsc{entropy} alongside high \textsc{Conc} and \textsc{HeadDiv}, indicating that most heads place almost all probability mass on a narrow set of mid-sequence tokens. This behavior might stem from its monotonically increasing denominators (seeEq.~\ref{eq:recurrent_nonapprox}). This could make it progressively harder for later tokens in the sequence to attract attention, thereby hindering effective training in uniform configurations. \textsl{StaticEmbQK} relaxing \texttt{Current QK} coupling, generally presents active token mixing from Layer 7, however, its mid-layers exhibit high similarity. Its reliance on static embeddings for attention computation limits its adaptability to individual layers, further constraining  predictive performance. \textsl{Approximate} and \textsl{FixedSeqQK}, showing attention patterns most similar to \textsl{Standard} across all layers. However, the performance of \textsl{FixedSeqQK} generally lags behind \textsl{Approximate}. This due to \textsl{FixedSeqQK}'s derivation of $\mathbf{Q}$ and $\mathbf{K}$ matrices from a fixed, pre-defined text sequence, which remains constant for all inputs. Consequently, the model might become prone to simulating this specific text sequence, thereby compromising its generalization ability. \textsl{RndEmbQK} attention faces a similar issue to \textsl{FixedSeqQK}, but suffers additional marginal performance drops, perhaps due to its inability to encode syntactic information.

\paragraph{Configurations.}

To illustrate the impact of different configurations, Fig.~\ref{fig:radar_24} shows the attention patterns of \textsl{RndEmbQK} and \textsl{Non-approximate} variants as representative methods for studying the behavior of different attention variants in \textit{uniform} and \textit{hybrid} configurations (see Fig.~\ref{fig:radar_hybrid_all_layers} for all layers).

With \textit{uniform} \textsl{RndEmbQK} (and \textit{uniform} \textsl{Standard}), the top-most layers (e.g. layer 22) exhibit high concentration (low \textsc{Entropy} and high \textsc{Conc}). This indicates a probability mass predominated by few selective tokens. In the hybrid design, those same layers become less selective (higher \textsc{Entropy}, lower \textsc{Conc}), leading to a decreased \textsc{Sink} score, suggesting that the hybrid mix alleviates first-token `sink' effects. In the \textsl{Non-approximate} hybrid model, odd layers keep the \textsl{Non-approximate} heads while even layers revert to \textsl{Standard}. A clear division of labor emerges: even (\textsl{Standard}) layers mirror the baseline, balancing token mixing and focus, while odd (\textsl{Non-approximate}) layers specialize, either acting as attention sinks (high \textsc{Sink}, low \textsc{Entropy}) or as mean-poolers (high \textsc{Entropy}, low \textsc{Conc}). This complementary interplay compensates for the lower expressiveness of \textsl{Non-approximate} heads observed in the uniform setting, explaining why the \textit{hybrid} configuration trains successfully while the \textit{uniform} one does not.

\paragraph{Why hybrid works.}

\begin{figure}[!t]
\centering
\includegraphics[width=\linewidth]{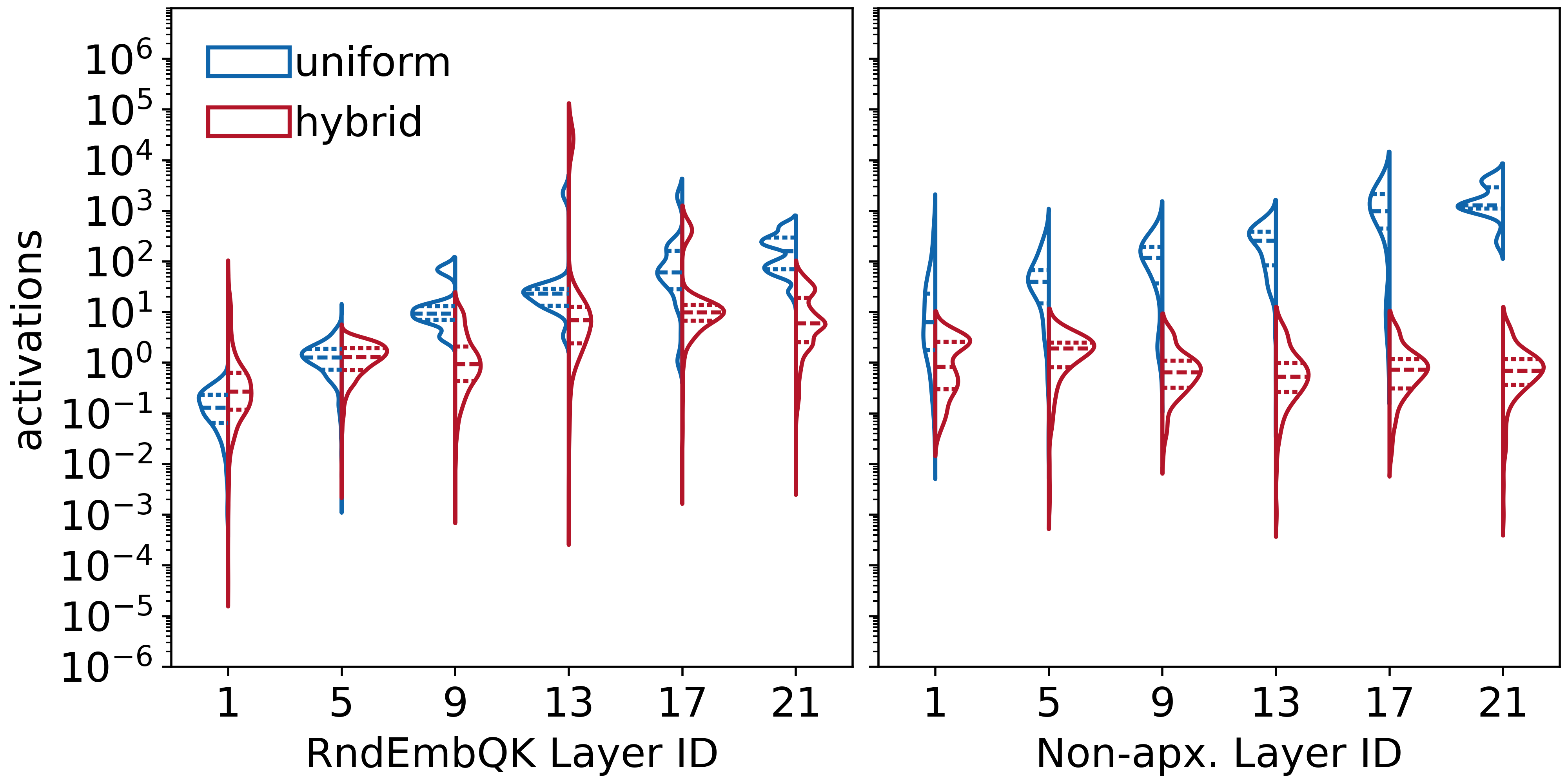}
\caption{Distribution of pre-softmax activations for \textsl{RndEmbQK} (left) and \textsl{Non-approximate} (right) across two different configurations. See Fig.~\ref{fig:violin_full} for all layers.}\label{fig:violin}
\end{figure}

We investigate the magnitude of raw activations (logits before softmax) within each \textsl{RndEmbQK} and \textsl{Non-approximate} layer in the \textit{hybrid} configuration (Fig.~\ref{fig:violin}). Our analysis reveals that activations generally exhibit lower magnitudes compared to the \textit{uniform} configuration for both attention variants. Notably, the \textit{uniform} \textsl{Non-approximate} model shows activation outliers exceeding $10^3$ in the final Transformer layers (e.g. Layer 21). In contrast, the \textit{hybrid} configuration maintains activations below $10^1$. This suggests that the \textsl{Standard} layers in the \textit{hybrid} architecture might serve as a normalization mechanism.  This normalization could mitigate over-concentration and the formation of highly sparse attention matrices, which can arise from large magnitude outliers during the numerically stable softmax operation. This normalizing effect appears sufficiently strong to rescue models that are otherwise challenging to train and prone to gradient vanishing (e.g. \textsl{Non-approximate} in the \textit{uniform} configuration).

\paragraph{Theoretical analysis.}

\citet{li2024spin} connects Transformer LMs to spin glass models. They suggest standard attention matrices align with the Gibbs-Boltzmann distribution~\citep{gibbs1902elementary}, implying an implicit energy minimization process with tokens as spins. Input-independent $\mathbf{Q}$ and $\mathbf{K}$ or form deviations disrupt this. This perspective provides a theoretical basis for the performance variations observed in our uniform replacement experiments. While \citet{zhang2022understanding} suggests full-rank attention offers maximal flexibility, causal attention can be low-rank due to stable softmax allowing zeros in diagonals with activation outliers. This supports our normalization analysis in \textit{hybrid} configurations, with \citet{neyshabur2017geometry}'s observation on unbalanced network training difficulty.

\paragraph{Model size.}
We also evaluate all attention variants across models of 70M, 160M, and 500M parameters. Our main observations remain consistent across these different model sizes. See Appx.~\ref{appendix:scaling_trend} for detailed results.

\paragraph{Hybrid configuration ablation.}\label{sec:different_hybrid_config}

To investigate the impact of replacing subsets of layers with simpler attention mechanisms, we consider nine different configurations. These focus on different segments of a 24-layer architecture of the 500M model: (1) \textit{even} or \textit{50\%} configuration, where even-numbered layers retain standard attention while odd-numbered layers are replaced; (2) \textit{odd} configuration, with the reverse arrangement; (3) \textit{top} configuration, where the upper layers (13-24) employ the simpler attention mechanism; (4) \textit{middle} configuration, targeting the middle layers (7-18); (5) \textit{bottom} configuration, focusing on the initial layers (1-6); (6) \textit{25\%}, replacing layers except Layer 4,8,12,16,20,24 with simpler attention; (7) \textit{first}, replacing all layers with simpler attention except the first layer; (8) \textit{last}, replacing all layers with simpler attention except the last layer; (9) \textit{bilateral}, replacing all layers with simpler attention except Layer 1 and 24. See Tbl.~\ref{table:model_config_abla} in Appx.~\ref{appendix:model_config_appendix_abla} for details.

\begin{figure}[!t]
\centering

\includegraphics[width=\linewidth]{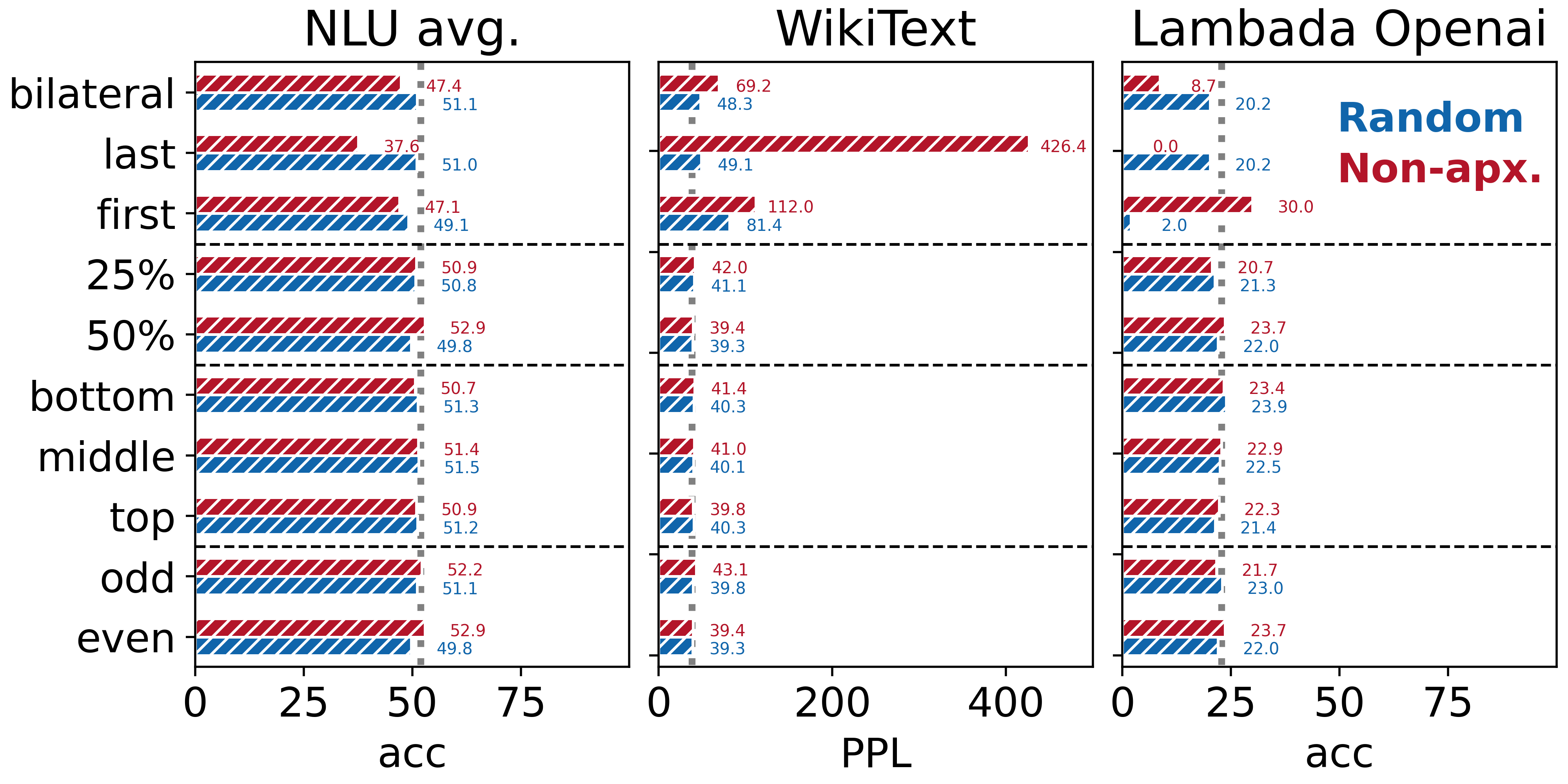}

\caption{Performance of \textsl{RndEmbQK} and \textsl{Non-approximate} across nine hybrid configurations. The vertical dotted lines represent the \textsl{Standard} baseline.}\label{fig:ablation}
\end{figure}

Fig.~\ref{fig:ablation} presents the predictive performance using these nine settings. For both \textsl{RndEmbQK} and \textsl{Non-approximate} mechanisms, the difference in performance across these hybrid configurations is marginal (e.g. all with a PPL around 40.0 on \textsc{WikiText}). However, this observation does not generalize to extreme settings, such as employing \textsl{Standard} attention in only the first or the last layer. For \textsl{RndEmbQK} attention, the predictive performance remains comparable to \textsl{Standard} if only the last layer (or layers at both ends) uses  \textsl{Standard}. Nevertheless, its accuracy on \textsc{Lambada Openai} drops to zero in such extreme cases. For \textsl{Non-approximate} attention, using \textsl{Standard} attention mechanism only in the last layer greatly harms performance, leading to PPL exceeding 400 on \textsc{WikiText}. This indicates that the normalization strength provided by a single \textsl{Standard} layer is limited. Therefore, in extreme hybrid settings where we can afford only one or two \textsl{Standard} layers, we should choose a substitute that still respects the main design principles presented in the \textit{uniform} setting (i.e. a stronger lightweight attention). Conversely, if the compute budget allows using even a small fraction of \textsl{Standard} transformer layers (e.g. 25\%), we can safely replace the remainder with a much simpler mechanism and still maintain competitive accuracy.

\section{Conclusion}

We systematically relax core design principles in a controlled setting, offering the first principled framework for assessing which aspects of attention are truly foundational and which can be safely simplified in language modeling. Our findings reveal that adhering to standard attention design principles varies between \textit{uniform} and \textit{hybrid} architectures. Token mixing and following the mathematical form are crucial for attention alternatives when applied uniformly, but not necessary for \textit{hybrid}. Strategically integrating a few standard attention layers within LMs can greatly improve, even overcome, limitations of less powerful attention mechanisms. This is likely due to the inherent normalization of standard attention, fostering training stability.

\section*{Limitations}

We performed experiments using a maximum model size of 500M parameters and a pretraining budget of 15B tokens, using a monolingual tokenizer and vocabulary, similar to \citet{allal2025smollm2,poli2023hyena}. While experimenting with larger models and different model families presents interesting avenues for future work, we believe that the current scope sufficiently supports our conclusions regarding the relative effectiveness of different attention designs.

\section*{Acknowledgments}

This projected made use of time on UK Tier-2 HPC facility JADE@ARC, funded by EPSRC (EP/T022205/1). We would like to thank Miles Williams and Atsuki Yamaguchi for their invaluable feedback.


\bibliography{anthology,custom}

\begin{thebibliography}{78}
\providecommand{\natexlab}[1]{#1}

\bibitem[{Ainslie et~al.(2023)Ainslie, Lee-Thorp, de~Jong, Zemlyanskiy, Lebron, and Sanghai}]{ainslie-etal-2023-gqa}
Joshua Ainslie, James Lee-Thorp, Michiel de~Jong, Yury Zemlyanskiy, Federico Lebron, and Sumit Sanghai. 2023.
\newblock \href {https://doi.org/10.18653/v1/2023.emnlp-main.298} {{GQA}: Training generalized multi-query transformer models from multi-head checkpoints}.
\newblock In \emph{Proceedings of the 2023 Conference on Empirical Methods in Natural Language Processing}, pages 4895--4901, Singapore. Association for Computational Linguistics.

\bibitem[{Aksenov et~al.(2024)Aksenov, Balagansky, Lo~Cicero~Vaina, Shaposhnikov, Gorbatovski, and Gavrilov}]{aksenov-etal-2024-linear}
Yaroslav Aksenov, Nikita Balagansky, Sofia Lo~Cicero~Vaina, Boris Shaposhnikov, Alexey Gorbatovski, and Daniil Gavrilov. 2024.
\newblock \href {https://doi.org/10.18653/v1/2024.acl-long.518} {Linear transformers with learnable kernel functions are better in-context models}.
\newblock In \emph{Proceedings of the 62nd Annual Meeting of the Association for Computational Linguistics (Volume 1: Long Papers)}, pages 9584--9597, Bangkok, Thailand. Association for Computational Linguistics.

\bibitem[{Allal et~al.(2025)Allal, Lozhkov, Bakouch, Bl{\'a}zquez, Penedo, Tunstall, Marafioti, Kydl{\'\i}cek, Lajar{\'\i}n, Srivastav et~al.}]{allal2025smollm2}
Loubna~Ben Allal, Anton Lozhkov, Elie Bakouch, Gabriel~Mart{\'\i}n Bl{\'a}zquez, Guilherme Penedo, Lewis Tunstall, Andr{\'e}s Marafioti, Hynek Kydl{\'\i}cek, Agust{\'\i}n~Piqueres Lajar{\'\i}n, Vaibhav Srivastav, and 1 others. 2025.
\newblock Smollm2: When smol goes big-data-centric training of a small language model.
\newblock \emph{CoRR}.

\bibitem[{Arora et~al.(2024)Arora, Eyuboglu, Zhang, Timalsina, Alberti, Zou, Rudra, and Re}]{arora2024simple}
Simran Arora, Sabri Eyuboglu, Michael Zhang, Aman Timalsina, Silas Alberti, James Zou, Atri Rudra, and Christopher Re. 2024.
\newblock \href {https://proceedings.mlr.press/v235/arora24a.html} {Simple linear attention language models balance the recall-throughput tradeoff}.
\newblock In \emph{Proceedings of the 41st International Conference on Machine Learning}, volume 235 of \emph{Proceedings of Machine Learning Research}, pages 1763--1840. PMLR.

\bibitem[{Baker(2022)}]{sep-simplicity}
Alan Baker. 2022.
\newblock {Simplicity}.
\newblock In Edward~N. Zalta, editor, \emph{The {Stanford} Encyclopedia of Philosophy}, {S}ummer 2022 edition. Metaphysics Research Lab, Stanford University.

\bibitem[{Behrouz et~al.(2024)Behrouz, Zhong, and Mirrokni}]{behrouz2024titans}
Ali Behrouz, Peilin Zhong, and Vahab Mirrokni. 2024.
\newblock {Titans: Learning to memorize at test time}.
\newblock \emph{arXiv preprint arXiv:2501.00663}.

\bibitem[{Biderman et~al.(2023)Biderman, Schoelkopf, Anthony, Bradley, O’Brien, Hallahan, Khan, Purohit, Prashanth, Raff et~al.}]{biderman2023pythia}
Stella Biderman, Hailey Schoelkopf, Quentin~Gregory Anthony, Herbie Bradley, Kyle O’Brien, Eric Hallahan, Mohammad~Aflah Khan, Shivanshu Purohit, USVSN~Sai Prashanth, Edward Raff, and 1 others. 2023.
\newblock {Pythia: A suite for analyzing large language models across training and scaling}.
\newblock In \emph{International Conference on Machine Learning}, pages 2397--2430. PMLR.

\bibitem[{Bisk et~al.(2020)Bisk, Zellers, Gao, Choi et~al.}]{bisk2020piqa}
Yonatan Bisk, Rowan Zellers, Jianfeng Gao, Yejin Choi, and 1 others. 2020.
\newblock {PIQA: Reasoning about Physical Commonsense in Natural Language}.
\newblock In \emph{Proceedings of the AAAI Conference on Artificial Intelligence}, volume~34, pages 7432--7439.

\bibitem[{Choromanski et~al.(2021)Choromanski, Likhosherstov, Dohan, Song, Gane, Sarlos, Hawkins, Davis, Mohiuddin, Kaiser, Belanger, Colwell, and Weller}]{choromanski2021rethinking}
Krzysztof~Marcin Choromanski, Valerii Likhosherstov, David Dohan, Xingyou Song, Andreea Gane, Tamas Sarlos, Peter Hawkins, Jared~Quincy Davis, Afroz Mohiuddin, Lukasz Kaiser, David~Benjamin Belanger, Lucy~J Colwell, and Adrian Weller. 2021.
\newblock \href {https://openreview.net/forum?id=Ua6zuk0WRH} {Rethinking attention with performers}.
\newblock In \emph{International Conference on Learning Representations}.

\bibitem[{Clark et~al.(2019)Clark, Lee, Chang, Kwiatkowski, Collins, and Toutanova}]{clark-etal-2019-boolq}
Christopher Clark, Kenton Lee, Ming-Wei Chang, Tom Kwiatkowski, Michael Collins, and Kristina Toutanova. 2019.
\newblock \href {https://doi.org/10.18653/v1/N19-1300} {{B}ool{Q}: Exploring the surprising difficulty of natural yes/no questions}.
\newblock In \emph{Proceedings of the 2019 Conference of the North {A}merican Chapter of the Association for Computational Linguistics: Human Language Technologies, Volume 1 (Long and Short Papers)}, pages 2924--2936, Minneapolis, Minnesota. Association for Computational Linguistics.

\bibitem[{Clark et~al.(2022)Clark, Garrette, Turc, and Wieting}]{clark-etal-2022-canine}
Jonathan~H. Clark, Dan Garrette, Iulia Turc, and John Wieting. 2022.
\newblock \href {https://doi.org/10.1162/tacl_a_00448} {Canine: Pre-training an efficient tokenization-free encoder for language representation}.
\newblock \emph{Transactions of the Association for Computational Linguistics}, 10:73--91.

\bibitem[{Clark et~al.(2018)Clark, Cowhey, Etzioni, Khot, Sabharwal, Schoenick, and Tafjord}]{clark2018think}
Peter Clark, Isaac Cowhey, Oren Etzioni, Tushar Khot, Ashish Sabharwal, Carissa Schoenick, and Oyvind Tafjord. 2018.
\newblock {Think you have solved question answering? Try arc, the AI2 reasoning challenge}.
\newblock \emph{arXiv preprint arXiv:1803.05457}.

\bibitem[{Dao and Gu(2024)}]{dao2024transformers}
Tri Dao and Albert Gu. 2024.
\newblock {Transformers are SSMs: generalized models and efficient algorithms through structured state space duality}.
\newblock In \emph{Proceedings of the 41st International Conference on Machine Learning}, pages 10041--10071.

\bibitem[{Devlin et~al.(2019)Devlin, Chang, Lee, and Toutanova}]{devlin-etal-2019-bert}
Jacob Devlin, Ming-Wei Chang, Kenton Lee, and Kristina Toutanova. 2019.
\newblock \href {https://doi.org/10.18653/v1/N19-1423} {{BERT}: Pre-training of deep bidirectional transformers for language understanding}.
\newblock In \emph{Proceedings of the 2019 Conference of the North {A}merican Chapter of the Association for Computational Linguistics: Human Language Technologies, Volume 1 (Long and Short Papers)}, pages 4171--4186, Minneapolis, Minnesota. Association for Computational Linguistics.

\bibitem[{Dong et~al.(2025)Dong, Fu, Diao, Byeon, CHEN, Mahabaleshwarkar, Liu, keirsbilck, Chen, Suhara, Lin, Kautz, and Molchanov}]{dong2025hymba}
Xin Dong, Yonggan Fu, Shizhe Diao, Wonmin Byeon, ZIJIA CHEN, Ameya~Sunil Mahabaleshwarkar, Shih-Yang Liu, Matthijs~Van keirsbilck, Min-Hung Chen, Yoshi Suhara, Yingyan~Celine Lin, Jan Kautz, and Pavlo Molchanov. 2025.
\newblock \href {https://openreview.net/forum?id=A1ztozypga} {Hymba: A hybrid-head architecture for small language models}.
\newblock In \emph{The Thirteenth International Conference on Learning Representations}.

\bibitem[{Elfwing et~al.(2018)Elfwing, Uchibe, and Doya}]{elfwing2018sigmoid}
Stefan Elfwing, Eiji Uchibe, and Kenji Doya. 2018.
\newblock Sigmoid-weighted linear units for neural network function approximation in reinforcement learning.
\newblock \emph{Neural networks}, 107:3--11.

\bibitem[{Fusco et~al.(2023)Fusco, Pascual, Staar, and Antognini}]{fusco-etal-2023-pnlp}
Francesco Fusco, Damian Pascual, Peter Staar, and Diego Antognini. 2023.
\newblock \href {https://doi.org/10.18653/v1/2023.acl-industry.6} {p{NLP}-mixer: an efficient all-{MLP} architecture for language}.
\newblock In \emph{Proceedings of the 61st Annual Meeting of the Association for Computational Linguistics (Volume 5: Industry Track)}, pages 53--60, Toronto, Canada. Association for Computational Linguistics.

\bibitem[{Gao et~al.(2024)Gao, Tow, Abbasi, Biderman, Black, DiPofi, Foster, Golding, Hsu, Le~Noac'h, Li, McDonell, Muennighoff, Ociepa, Phang, Reynolds, Schoelkopf, Skowron, Sutawika, Tang, Thite, Wang, Wang, and Zou}]{eval-harness}
Leo Gao, Jonathan Tow, Baber Abbasi, Stella Biderman, Sid Black, Anthony DiPofi, Charles Foster, Laurence Golding, Jeffrey Hsu, Alain Le~Noac'h, Haonan Li, Kyle McDonell, Niklas Muennighoff, Chris Ociepa, Jason Phang, Laria Reynolds, Hailey Schoelkopf, Aviya Skowron, Lintang Sutawika, and 5 others. 2024.
\newblock \href {https://doi.org/10.5281/zenodo.12608602} {A framework for few-shot language model evaluation}.

\bibitem[{Gibbs(1902)}]{gibbs1902elementary}
Josiah~Willard Gibbs. 1902.
\newblock \emph{{Elementary principles in statistical mechanics: Developed with especial reference to the rational foundations of thermodynamics}}.
\newblock C. Scribner's sons.

\bibitem[{Glorioso et~al.(2024)Glorioso, Anthony, Tokpanov, Whittington, Pilault, Ibrahim, and Millidge}]{glorioso2024zamba}
Paolo Glorioso, Quentin Anthony, Yury Tokpanov, James Whittington, Jonathan Pilault, Adam Ibrahim, and Beren Millidge. 2024.
\newblock {Zamba: A compact 7B SSM hybrid model}.
\newblock \emph{arXiv preprint arXiv:2405.16712}.

\bibitem[{Gu and Dao(2024)}]{gu2024mamba}
Albert Gu and Tri Dao. 2024.
\newblock \href {https://openreview.net/forum?id=tEYskw1VY2} {Mamba: Linear-time sequence modeling with selective state spaces}.
\newblock In \emph{First Conference on Language Modeling}.

\bibitem[{Gu et~al.(2022)Gu, Goel, and Re}]{gu2022efficiently}
Albert Gu, Karan Goel, and Christopher Re. 2022.
\newblock \href {https://openreview.net/forum?id=uYLFoz1vlAC} {Efficiently modeling long sequences with structured state spaces}.
\newblock In \emph{International Conference on Learning Representations}.

\bibitem[{Han et~al.(2024)Han, Wang, Peng, Xiong, Chen, Ji, and Wang}]{han-etal-2024-lm}
Chi Han, Qifan Wang, Hao Peng, Wenhan Xiong, Yu~Chen, Heng Ji, and Sinong Wang. 2024.
\newblock \href {https://doi.org/10.18653/v1/2024.naacl-long.222} {{LM}-infinite: Zero-shot extreme length generalization for large language models}.
\newblock In \emph{Proceedings of the 2024 Conference of the North American Chapter of the Association for Computational Linguistics: Human Language Technologies (Volume 1: Long Papers)}, pages 3991--4008, Mexico City, Mexico. Association for Computational Linguistics.

\bibitem[{He et~al.(2025)He, Yu, Gong, Liu, Li, and Lin}]{he2025rodimus}
Zhihao He, Hang Yu, Zi~Gong, Shizhan Liu, Jianguo Li, and Weiyao Lin. 2025.
\newblock \href {https://openreview.net/forum?id=IIVYiJ1ggK} {Rodimus*: Breaking the accuracy-efficiency trade-off with efficient attentions}.
\newblock In \emph{The Thirteenth International Conference on Learning Representations}.

\bibitem[{Hoffmann et~al.(2022)Hoffmann, Borgeaud, Mensch, Buchatskaya, Cai, Rutherford, de~Las~Casas, Hendricks, Welbl, Clark, Hennigan, Noland, Millican, van~den Driessche, Damoc, Guy, Osindero, Simonyan, Elsen, Vinyals, Rae, and Sifre}]{hoffmann2022training}
Jordan Hoffmann, Sebastian Borgeaud, Arthur Mensch, Elena Buchatskaya, Trevor Cai, Eliza Rutherford, Diego de~Las~Casas, Lisa~Anne Hendricks, Johannes Welbl, Aidan Clark, Tom Hennigan, Eric Noland, Katie Millican, George van~den Driessche, Bogdan Damoc, Aurelia Guy, Simon Osindero, Karen Simonyan, Erich Elsen, and 3 others. 2022.
\newblock Training compute-optimal large language models.
\newblock In \emph{Proceedings of the 36th International Conference on Neural Information Processing Systems}, NIPS '22, Red Hook, NY, USA. Curran Associates Inc.

\bibitem[{Kasai et~al.(2021)Kasai, Peng, Zhang, Yogatama, Ilharco, Pappas, Mao, Chen, and Smith}]{kasai-etal-2021-finetuning}
Jungo Kasai, Hao Peng, Yizhe Zhang, Dani Yogatama, Gabriel Ilharco, Nikolaos Pappas, Yi~Mao, Weizhu Chen, and Noah~A. Smith. 2021.
\newblock \href {https://doi.org/10.18653/v1/2021.emnlp-main.830} {Finetuning pretrained transformers into {RNN}s}.
\newblock In \emph{Proceedings of the 2021 Conference on Empirical Methods in Natural Language Processing}, pages 10630--10643, Online and Punta Cana, Dominican Republic. Association for Computational Linguistics.

\bibitem[{Katharopoulos et~al.(2020)Katharopoulos, Vyas, Pappas, and Fleuret}]{katharopoulos2020transformers}
Angelos Katharopoulos, Apoorv Vyas, Nikolaos Pappas, and Fran{\c{c}}ois Fleuret. 2020.
\newblock {Transformers are RNNs: Fast autoregressive transformers with linear attention}.
\newblock In \emph{International Conference on Machine Learning}, pages 5156--5165. PMLR.

\bibitem[{Kitaev et~al.(2020)Kitaev, Kaiser, and Levskaya}]{Kitaev2020Reformer:}
Nikita Kitaev, Lukasz Kaiser, and Anselm Levskaya. 2020.
\newblock \href {https://openreview.net/forum?id=rkgNKkHtvB} {Reformer: The efficient transformer}.
\newblock In \emph{International Conference on Learning Representations}.

\bibitem[{Korthikanti et~al.(2023)Korthikanti, Casper, Lym, McAfee, Andersch, Shoeybi, and Catanzaro}]{korthikanti2023reducing}
Vijay~Anand Korthikanti, Jared Casper, Sangkug Lym, Lawrence McAfee, Michael Andersch, Mohammad Shoeybi, and Bryan Catanzaro. 2023.
\newblock Reducing activation recomputation in large transformer models.
\newblock \emph{Proceedings of Machine Learning and Systems}, 5:341--353.

\bibitem[{Lee-Thorp et~al.(2022)Lee-Thorp, Ainslie, Eckstein, and Ontanon}]{lee-thorp-etal-2022-fnet}
James Lee-Thorp, Joshua Ainslie, Ilya Eckstein, and Santiago Ontanon. 2022.
\newblock \href {https://doi.org/10.18653/v1/2022.naacl-main.319} {{FN}et: Mixing tokens with {F}ourier transforms}.
\newblock In \emph{Proceedings of the 2022 Conference of the North American Chapter of the Association for Computational Linguistics: Human Language Technologies}, pages 4296--4313, Seattle, United States. Association for Computational Linguistics.

\bibitem[{Lenz et~al.(2025)Lenz, Lieber, Arazi, Bergman, Manevich, Peleg, Aviram, Almagor, Fridman, Padnos, Gissin, Jannai, Muhlgay, Zimberg, Gerber, Dolev, Krakovsky, Safahi, Schwartz, Cohen, Shachaf, Rozenblum, Bata, Blass, Magar, Dalmedigos, Osin, Fadlon, Rozman, Danos, Gokhman, Zusman, Gidron, Ratner, Gat, Rozen, Fried, Leshno, Antverg, Abend, Dagan, Cohavi, Alon, Belson, Cohen, Gilad, Glozman, Lev, Shalev-Shwartz, Meirom, Delbari, Ness, Asida, Gal, Braude, Pumerantz, Cohen, Belinkov, Globerson, Levy, and Shoham}]{lenz2025jamba}
Barak Lenz, Opher Lieber, Alan Arazi, Amir Bergman, Avshalom Manevich, Barak Peleg, Ben Aviram, Chen Almagor, Clara Fridman, Dan Padnos, Daniel Gissin, Daniel Jannai, Dor Muhlgay, Dor Zimberg, Edden~M. Gerber, Elad Dolev, Eran Krakovsky, Erez Safahi, Erez Schwartz, and 42 others. 2025.
\newblock \href {https://openreview.net/forum?id=JFPaD7lpBD} {{Jamba: Hybrid Transformer-Mamba language models}}.
\newblock In \emph{The Thirteenth International Conference on Learning Representations}.

\bibitem[{Li et~al.(2024)Li, Bai, and Huang}]{li2024spin}
Yuhao Li, Ruoran Bai, and Haiping Huang. 2024.
\newblock Spin glass model of in-context learning.
\newblock \emph{arXiv preprint arXiv:2408.02288}.

\bibitem[{Lin et~al.(2025)Lin, Nikishin, He, and Courville}]{lin2025forgetting}
Zhixuan Lin, Evgenii Nikishin, Xu~He, and Aaron Courville. 2025.
\newblock \href {https://openreview.net/forum?id=q2Lnyegkr8} {Forgetting transformer: Softmax attention with a forget gate}.
\newblock In \emph{The Thirteenth International Conference on Learning Representations}.

\bibitem[{Liu et~al.(2024)Liu, Feng, Xue, Wang, Wu, Lu, Zhao, Deng, Zhang, Ruan et~al.}]{liu2024deepseek}
Aixin Liu, Bei Feng, Bing Xue, Bingxuan Wang, Bochao Wu, Chengda Lu, Chenggang Zhao, Chengqi Deng, Chenyu Zhang, Chong Ruan, and 1 others. 2024.
\newblock Deepseek-v3 technical report.
\newblock \emph{arXiv preprint arXiv:2412.19437}.

\bibitem[{Lozhkov et~al.(2024)Lozhkov, Ben~Allal, von Werra, and Wolf}]{lozhkov2024fineweb-edu}
Anton Lozhkov, Loubna Ben~Allal, Leandro von Werra, and Thomas Wolf. 2024.
\newblock \href {https://doi.org/10.57967/hf/2497} {{FineWeb-Edu: The finest collection of educational content }}.

\bibitem[{Merity et~al.(2017)Merity, Xiong, Bradbury, and Socher}]{merity2017pointer}
Stephen Merity, Caiming Xiong, James Bradbury, and Richard Socher. 2017.
\newblock \href {https://openreview.net/forum?id=Byj72udxe} {Pointer sentinel mixture models}.
\newblock In \emph{International Conference on Learning Representations}.

\bibitem[{Narayanan et~al.(2021)Narayanan, Shoeybi, Casper, LeGresley, Patwary, Korthikanti, Vainbrand, Kashinkunti, Bernauer, Catanzaro et~al.}]{narayanan2021efficient}
Deepak Narayanan, Mohammad Shoeybi, Jared Casper, Patrick LeGresley, Mostofa Patwary, Vijay Korthikanti, Dmitri Vainbrand, Prethvi Kashinkunti, Julie Bernauer, Bryan Catanzaro, and 1 others. 2021.
\newblock Efficient large-scale language model training on gpu clusters using megatron-lm.
\newblock In \emph{Proceedings of the international conference for high performance computing, networking, storage and analysis}, pages 1--15.

\bibitem[{Nawrot et~al.(2023)Nawrot, Chorowski, Lancucki, and Ponti}]{nawrot-etal-2023-efficient}
Piotr Nawrot, Jan Chorowski, Adrian Lancucki, and Edoardo~Maria Ponti. 2023.
\newblock \href {https://doi.org/10.18653/v1/2023.acl-long.353} {Efficient transformers with dynamic token pooling}.
\newblock In \emph{Proceedings of the 61st Annual Meeting of the Association for Computational Linguistics (Volume 1: Long Papers)}, pages 6403--6417, Toronto, Canada. Association for Computational Linguistics.

\bibitem[{Neyshabur et~al.(2017)Neyshabur, Tomioka, Salakhutdinov, and Srebro}]{neyshabur2017geometry}
Behnam Neyshabur, Ryota Tomioka, Ruslan Salakhutdinov, and Nathan Srebro. 2017.
\newblock Geometry of optimization and implicit regularization in deep learning.
\newblock \emph{arXiv preprint arXiv:1705.03071}.

\bibitem[{Orvieto et~al.(2023)Orvieto, Smith, Gu, Fernando, Gulcehre, Pascanu, and De}]{orvieto2023resurrecting}
Antonio Orvieto, Samuel~L Smith, Albert Gu, Anushan Fernando, Caglar Gulcehre, Razvan Pascanu, and Soham De. 2023.
\newblock Resurrecting recurrent neural networks for long sequences.
\newblock In \emph{International Conference on Machine Learning}, pages 26670--26698. PMLR.

\bibitem[{Peng et~al.(2023)Peng, Alcaide, Anthony, Albalak, Arcadinho, Biderman, Cao, Cheng, Chung, Derczynski, Du, Grella, Gv, He, Hou, Kazienko, Kocon, Kong, Koptyra, Lau, Lin, Mantri, Mom, Saito, Song, Tang, Wind, Wo{\'z}niak, Zhang, Zhou, Zhu, and Zhu}]{peng-etal-2023-rwkv}
Bo~Peng, Eric Alcaide, Quentin Anthony, Alon Albalak, Samuel Arcadinho, Stella Biderman, Huanqi Cao, Xin Cheng, Michael Chung, Leon Derczynski, Xingjian Du, Matteo Grella, Kranthi Gv, Xuzheng He, Haowen Hou, Przemyslaw Kazienko, Jan Kocon, Jiaming Kong, Bart{\l}omiej Koptyra, and 13 others. 2023.
\newblock \href {https://doi.org/10.18653/v1/2023.findings-emnlp.936} {{RWKV}: Reinventing {RNN}s for the transformer era}.
\newblock In \emph{Findings of the Association for Computational Linguistics: EMNLP 2023}, pages 14048--14077, Singapore. Association for Computational Linguistics.

\bibitem[{Peng et~al.(2024)Peng, Goldstein, Anthony, Albalak, Alcaide, Biderman, Cheah, Ferdinan, GV, Hou, Krishna, Jr., Muennighoff, Obeid, Saito, Song, Tu, Zhang, Zhao, Zhao, Zhu, and Zhu}]{peng2024eagle}
Bo~Peng, Daniel Goldstein, Quentin~Gregory Anthony, Alon Albalak, Eric Alcaide, Stella Biderman, Eugene Cheah, Teddy Ferdinan, Kranthi~Kiran GV, Haowen Hou, Satyapriya Krishna, Ronald~McClelland Jr., Niklas Muennighoff, Fares Obeid, Atsushi Saito, Guangyu Song, Haoqin Tu, Ruichong Zhang, Bingchen Zhao, and 3 others. 2024.
\newblock \href {https://openreview.net/forum?id=soz1SEiPeq} {{Eagle and Finch: RWKV with matrix-valued states and dynamic recurrence}}.
\newblock In \emph{First Conference on Language Modeling}.

\bibitem[{Peng et~al.(2025)Peng, Zhang, Goldstein, Alcaide, Hou, Lu, Merrill, Song, Tan, Utpala et~al.}]{peng2025rwkv}
Bo~Peng, Ruichong Zhang, Daniel Goldstein, Eric Alcaide, Haowen Hou, Janna Lu, William Merrill, Guangyu Song, Kaifeng Tan, Saiteja Utpala, and 1 others. 2025.
\newblock {RWKV-7" Goose" with expressive dynamic state evolution}.
\newblock \emph{arXiv preprint arXiv:2503.14456}.

\bibitem[{Peng and Cao(2024)}]{peng2024etamba}
Dazhi Peng and Hangrui Cao. 2024.
\newblock \href {https://openreview.net/forum?id=cbuBBgOHq8} {{E-Tamba: Efficient Transformer-Mamba layer transplantation}}.
\newblock In \emph{NeurIPS 2024 Workshop on Fine-Tuning in Modern Machine Learning: Principles and Scalability}.

\bibitem[{Peng et~al.(2021)Peng, Pappas, Yogatama, Schwartz, Smith, and Kong}]{peng2021random}
Hao Peng, Nikolaos Pappas, Dani Yogatama, Roy Schwartz, Noah Smith, and Lingpeng Kong. 2021.
\newblock \href {https://openreview.net/forum?id=QtTKTdVrFBB} {Random feature attention}.
\newblock In \emph{International Conference on Learning Representations}.

\bibitem[{Poli et~al.(2023)Poli, Massaroli, Nguyen, Fu, Dao, Baccus, Bengio, Ermon, and R{\'e}}]{poli2023hyena}
Michael Poli, Stefano Massaroli, Eric Nguyen, Daniel~Y Fu, Tri Dao, Stephen Baccus, Yoshua Bengio, Stefano Ermon, and Christopher R{\'e}. 2023.
\newblock {Hyena hierarchy: Towards larger convolutional language models}.
\newblock In \emph{International Conference on Machine Learning}, pages 28043--28078. PMLR.

\bibitem[{Qin et~al.(2022)Qin, Sun, Deng, Li, Wei, Lv, Yan, Kong, and Zhong}]{zhen2022cosformer}
Zhen Qin, Weixuan Sun, Hui Deng, Dongxu Li, Yunshen Wei, Baohong Lv, Junjie Yan, Lingpeng Kong, and Yiran Zhong. 2022.
\newblock \href {https://openreview.net/forum?id=Bl8CQrx2Up4} {{cosFormer: Rethinking Softmax In Attention}}.
\newblock In \emph{International Conference on Learning Representations}.

\bibitem[{Qin et~al.(2024)Qin, Yang, Sun, Shen, Li, Sun, and Zhong}]{qin2024hgrn}
Zhen Qin, Songlin Yang, Weixuan Sun, Xuyang Shen, Dong Li, Weigao Sun, and Yiran Zhong. 2024.
\newblock \href {https://openreview.net/forum?id=y6SqbJfCSk} {{HGRN2: Gated linear RNNs with state expansion}}.
\newblock In \emph{First Conference on Language Modeling}.

\bibitem[{Radford et~al.(2019)Radford, Wu, Child, Luan, Amodei, and Sutskever}]{radford2019language}
Alec Radford, Jeff Wu, Rewon Child, David Luan, Dario Amodei, and Ilya Sutskever. 2019.
\newblock Language models are unsupervised multitask learners.

\bibitem[{Rajabzadeh et~al.(2024)Rajabzadeh, Jafari, Sharma, Jami, Kwon, Ghodsi, Chen, and Rezagholizadeh}]{rajabzadeh2024echoatt}
Hossein Rajabzadeh, Aref Jafari, Aman Sharma, Benyamin Jami, Hyock Ju~Hj Kwon, Ali Ghodsi, Boxing Chen, and Mehdi Rezagholizadeh. 2024.
\newblock Echoatt: Attend, copy, then adjust for more efficient large language models.
\newblock In \emph{NeurIPS Efficient Natural Language and Speech Processing Workshop}, pages 259--269. PMLR.

\bibitem[{Ren et~al.(2025)Ren, Ma, Yang, Wei, Zhang, and Chen}]{ren2025vamba}
Weiming Ren, Wentao Ma, Huan Yang, Cong Wei, Ge~Zhang, and Wenhu Chen. 2025.
\newblock {Vamba: Understanding hour-long videos with hybrid mamba-transformers}.
\newblock \emph{arXiv preprint arXiv:2503.11579}.

\bibitem[{Roemmele et~al.(2011)Roemmele, Bejan, and Gordon}]{roemmele2011choice}
Melissa Roemmele, Cosmin~Adrian Bejan, and Andrew~S Gordon. 2011.
\newblock {Choice of Plausible Alternatives: An Evaluation of Commonsense Causal Reasoning.}
\newblock In \emph{AAAI spring symposium: logical formalizations of commonsense reasoning}, pages 90--95.

\bibitem[{Schlag et~al.(2021)Schlag, Irie, and Schmidhuber}]{schlag2021linear}
Imanol Schlag, Kazuki Irie, and J{\"u}rgen Schmidhuber. 2021.
\newblock {Linear transformers are secretly fast weight programmers}.
\newblock In \emph{International conference on machine learning}, pages 9355--9366. PMLR.

\bibitem[{Sennrich et~al.(2016)Sennrich, Haddow, and Birch}]{sennrich-etal-2016-neural}
Rico Sennrich, Barry Haddow, and Alexandra Birch. 2016.
\newblock \href {https://doi.org/10.18653/v1/P16-1162} {Neural machine translation of rare words with subword units}.
\newblock In \emph{Proceedings of the 54th Annual Meeting of the Association for Computational Linguistics (Volume 1: Long Papers)}, pages 1715--1725, Berlin, Germany. Association for Computational Linguistics.

\bibitem[{Shazeer(2019)}]{shazeer2019fast}
Noam Shazeer. 2019.
\newblock Fast transformer decoding: One write-head is all you need.
\newblock \emph{arXiv preprint arXiv:1911.02150}.

\bibitem[{Siems et~al.(2025)Siems, Carstensen, Zela, Hutter, Pontil, and Grazzi}]{siems2025deltaproduct}
Julien Siems, Timur Carstensen, Arber Zela, Frank Hutter, Massimiliano Pontil, and Riccardo Grazzi. 2025.
\newblock {DeltaProduct: Improving state-tracking in linear RNNs via Householder products}.
\newblock \emph{arXiv preprint arXiv:2502.10297}.

\bibitem[{Singh(2025)}]{singh2025meta}
Ajit Singh. 2025.
\newblock {Meta Llama 4: The future of multimodal AI}.
\newblock \emph{Available at SSRN 5208228}.

\bibitem[{Smith et~al.(2022)Smith, Patwary, Norick, LeGresley, Rajbhandari, Casper, Liu, Prabhumoye, Zerveas, Korthikanti et~al.}]{smith2022using}
Shaden Smith, Mostofa Patwary, Brandon Norick, Patrick LeGresley, Samyam Rajbhandari, Jared Casper, Zhun Liu, Shrimai Prabhumoye, George Zerveas, Vijay Korthikanti, and 1 others. 2022.
\newblock Using deepspeed and megatron to train megatron-turing nlg 530b, a large-scale generative language model.
\newblock \emph{arXiv preprint arXiv:2201.11990}.

\bibitem[{Soboleva et~al.(2023)Soboleva, Al-Khateeb, Myers, Steeves, Hestness, and Dey}]{cerebras2023slimpajama}
Daria Soboleva, Faisal Al-Khateeb, Robert Myers, Jacob~R Steeves, Joel Hestness, and Nolan Dey. 2023.
\newblock {SlimPajama: A 627B token cleaned and deduplicated version of RedPajama}.

\bibitem[{Tay et~al.(2022)Tay, Dehghani, Bahri, and Metzler}]{tay2022efficient}
Yi~Tay, Mostafa Dehghani, Dara Bahri, and Donald Metzler. 2022.
\newblock {Efficient transformers: A survey}.
\newblock \emph{ACM Computing Surveys}, 55(6):1--28.

\bibitem[{Tay et~al.(2019)Tay, Zhang, Luu, Rao, Zhang, Wang, Fu, and Hui}]{tay-etal-2019-lightweight}
Yi~Tay, Aston Zhang, Anh~Tuan Luu, Jinfeng Rao, Shuai Zhang, Shuohang Wang, Jie Fu, and Siu~Cheung Hui. 2019.
\newblock \href {https://doi.org/10.18653/v1/P19-1145} {Lightweight and efficient neural natural language processing with quaternion networks}.
\newblock In \emph{Proceedings of the 57th Annual Meeting of the Association for Computational Linguistics}, pages 1494--1503, Florence, Italy. Association for Computational Linguistics.

\bibitem[{Team et~al.(2024)Team, Lenz, Arazi, Bergman, Manevich, Peleg, Aviram, Almagor, Fridman, Padnos et~al.}]{team2024jamba}
Jamba Team, Barak Lenz, Alan Arazi, Amir Bergman, Avshalom Manevich, Barak Peleg, Ben Aviram, Chen Almagor, Clara Fridman, Dan Padnos, and 1 others. 2024.
\newblock {Jamba-1.5: Hybrid transformer-mamba models at scale}.
\newblock \emph{arXiv preprint arXiv:2408.12570}.

\bibitem[{Tolstikhin et~al.(2021)Tolstikhin, Houlsby, Kolesnikov, Beyer, Zhai, Unterthiner, Yung, Steiner, Keysers, Uszkoreit et~al.}]{tolstikhin2021mlp}
Ilya~O Tolstikhin, Neil Houlsby, Alexander Kolesnikov, Lucas Beyer, Xiaohua Zhai, Thomas Unterthiner, Jessica Yung, Andreas Steiner, Daniel Keysers, Jakob Uszkoreit, and 1 others. 2021.
\newblock {MLP-Mixer: An all-mlp architecture for vision}.
\newblock \emph{Advances in neural information processing systems}, 34:24261--24272.

\bibitem[{Vaswani et~al.(2017)Vaswani, Shazeer, Parmar, Uszkoreit, Jones, Gomez, Kaiser, and Polosukhin}]{vaswani2017attention}
Ashish Vaswani, Noam Shazeer, Niki Parmar, Jakob Uszkoreit, Llion Jones, Aidan~N Gomez, {\L}ukasz Kaiser, and Illia Polosukhin. 2017.
\newblock Attention is all you need.
\newblock \emph{Advances in neural information processing systems}, 30.

\bibitem[{Wang et~al.(2019)Wang, Pruksachatkun, Nangia, Singh, Michael, Hill, Levy, and Bowman}]{wang2019superglue}
Alex Wang, Yada Pruksachatkun, Nikita Nangia, Amanpreet Singh, Julian Michael, Felix Hill, Omer Levy, and Samuel Bowman. 2019.
\newblock {Superglue: A stickier benchmark for general-purpose language understanding systems}.
\newblock \emph{Advances in neural information processing systems}, 32.

\bibitem[{Wang et~al.(2020)Wang, Li, Khabsa, Fang, and Ma}]{wang2020linformer}
Sinong Wang, Belinda~Z Li, Madian Khabsa, Han Fang, and Hao Ma. 2020.
\newblock {Linformer: Self-attention with linear complexity}.
\newblock \emph{arXiv preprint arXiv:2006.04768}.

\bibitem[{Welbl et~al.(2017)Welbl, Liu, and Gardner}]{welbl-etal-2017-crowdsourcing}
Johannes Welbl, Nelson~F. Liu, and Matt Gardner. 2017.
\newblock \href {https://doi.org/10.18653/v1/W17-4413} {Crowdsourcing multiple choice science questions}.
\newblock In \emph{Proceedings of the 3rd Workshop on Noisy User-generated Text}, pages 94--106, Copenhagen, Denmark. Association for Computational Linguistics.

\bibitem[{Xiao et~al.(2024)Xiao, Tian, Chen, Han, and Lewis}]{xiao2024efficient}
Guangxuan Xiao, Yuandong Tian, Beidi Chen, Song Han, and Mike Lewis. 2024.
\newblock \href {https://openreview.net/forum?id=NG7sS51zVF} {Efficient streaming language models with attention sinks}.
\newblock In \emph{The Twelfth International Conference on Learning Representations}.

\bibitem[{Xiao et~al.(2019)Xiao, Li, Zhu, Yu, and Liu}]{xiao2019sharing}
Tong Xiao, Yinqiao Li, Jingbo Zhu, Zhengtao Yu, and Tongran Liu. 2019.
\newblock Sharing attention weights for fast transformer.
\newblock \emph{arXiv preprint arXiv:1906.11024}.

\bibitem[{Xue and Aletras(2022)}]{xue-aletras-2022-hashformers}
Huiyin Xue and Nikolaos Aletras. 2022.
\newblock \href {https://doi.org/10.18653/v1/2022.emnlp-main.536} {{H}ash{F}ormers: Towards vocabulary-independent pre-trained transformers}.
\newblock In \emph{Proceedings of the 2022 Conference on Empirical Methods in Natural Language Processing}, pages 7862--7874, Abu Dhabi, United Arab Emirates. Association for Computational Linguistics.

\bibitem[{Xue and Aletras(2023)}]{xue-aletras-2023-pit}
Huiyin Xue and Nikolaos Aletras. 2023.
\newblock \href {https://doi.org/10.18653/v1/2023.findings-emnlp.695} {Pit one against many: Leveraging attention-head embeddings for parameter-efficient multi-head attention}.
\newblock In \emph{Findings of the Association for Computational Linguistics: EMNLP 2023}, pages 10355--10373, Singapore. Association for Computational Linguistics.

\bibitem[{Yan et~al.(2021)Yan, Chen, Qi, Bhendawade, Gong, Duan, and Zhang}]{yan2021attention}
Yu~Yan, Jiusheng Chen, Weizhen Qi, Nikhil Bhendawade, Yeyun Gong, Nan Duan, and Ruofei Zhang. 2021.
\newblock El-attention: Memory efficient lossless attention for generation.
\newblock In \emph{International Conference on Machine Learning}, pages 11648--11658. PMLR.

\bibitem[{Yang et~al.(2025)Yang, Li, Yang, Zhang, Hui, Zheng, Yu, Gao, Huang, Lv et~al.}]{yang2025qwen3}
An~Yang, Anfeng Li, Baosong Yang, Beichen Zhang, Binyuan Hui, Bo~Zheng, Bowen Yu, Chang Gao, Chengen Huang, Chenxu Lv, and 1 others. 2025.
\newblock Qwen3 technical report.
\newblock \emph{arXiv preprint arXiv:2505.09388}.

\bibitem[{Yang et~al.(2024{\natexlab{a}})Yang, Yang, Zhang, Hui, Zheng, Yu, Li, Liu, Huang, Wei et~al.}]{yang2024qwen2}
An~Yang, Baosong Yang, Beichen Zhang, Binyuan Hui, Bo~Zheng, Bowen Yu, Chengyuan Li, Dayiheng Liu, Fei Huang, Haoran Wei, and 1 others. 2024{\natexlab{a}}.
\newblock Qwen2. 5 technical report.
\newblock \emph{arXiv preprint arXiv:2412.15115}.

\bibitem[{Yang et~al.(2024{\natexlab{b}})Yang, Wang, Zhang, Shen, and Kim}]{yang2024parallelizing}
Songlin Yang, Bailin Wang, Yu~Zhang, Yikang Shen, and Yoon Kim. 2024{\natexlab{b}}.
\newblock \href {https://openreview.net/forum?id=y8Rm4VNRPH} {Parallelizing linear transformers with the delta rule over sequence length}.
\newblock In \emph{The Thirty-eighth Annual Conference on Neural Information Processing Systems}.

\bibitem[{Yu et~al.(2022)Yu, Luo, Zhou, Si, Zhou, Wang, Feng, and Yan}]{yu2022metaformer}
Weihao Yu, Mi~Luo, Pan Zhou, Chenyang Si, Yichen Zhou, Xinchao Wang, Jiashi Feng, and Shuicheng Yan. 2022.
\newblock Metaformer is actually what you need for vision.
\newblock In \emph{Proceedings of the IEEE/CVF conference on computer vision and pattern recognition}, pages 10819--10829.

\bibitem[{Zellers et~al.(2019)Zellers, Holtzman, Bisk, Farhadi, and Choi}]{zellers-etal-2019-hellaswag}
Rowan Zellers, Ari Holtzman, Yonatan Bisk, Ali Farhadi, and Yejin Choi. 2019.
\newblock \href {https://doi.org/10.18653/v1/P19-1472} {{H}ella{S}wag: Can a machine really finish your sentence?}
\newblock In \emph{Proceedings of the 57th Annual Meeting of the Association for Computational Linguistics}, pages 4791--4800, Florence, Italy. Association for Computational Linguistics.

\bibitem[{Zhang et~al.(2022)Zhang, Bengio, Hardt, Recht, and Vinyals}]{zhang2022understanding}
Chiyuan Zhang, Samy Bengio, Moritz Hardt, Benjamin Recht, and Oriol Vinyals. 2022.
\newblock Understanding deep learning requires rethinking generalization.
\newblock In \emph{International Conference on Learning Representations}.

\end{thebibliography}

\appendix

\section{Experiments with Different Model Sizes}\label{appendix:scaling_trend}

\begin{figure}[!t]
\centering
{%
\includegraphics[width=\linewidth]{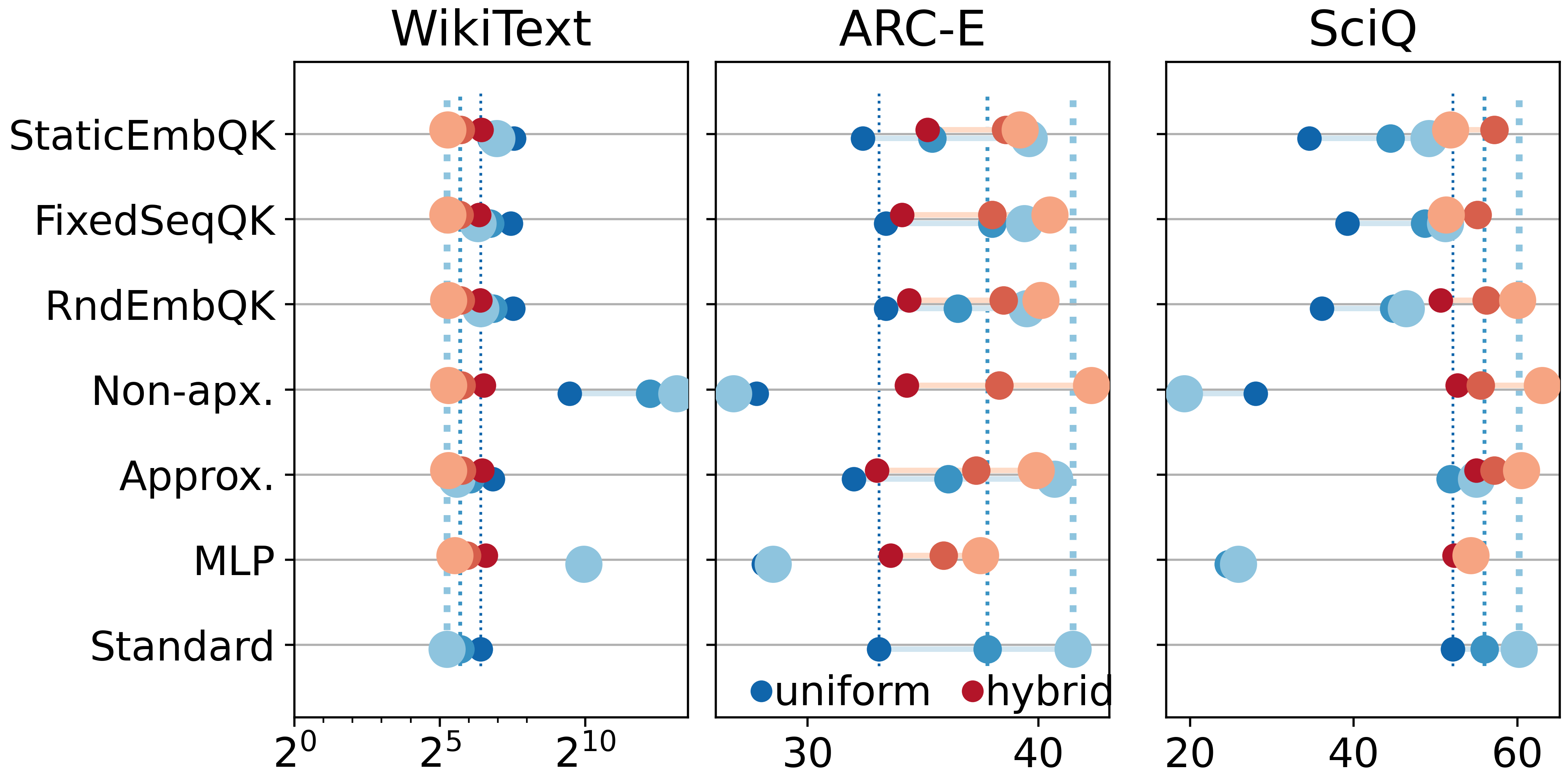}
}%
\caption{Predictive performance of 70M parameters (small dots), 160M parameters (medium dots), and 500M parameters (large dots) models with different attention mechanisms and configurations on \textsc{WikiText}, \textsc{ARC-E}, and \textsc{SciQ}.}\label{fig:scaling}
\end{figure}

To assess the impact of model size, we evaluate all attention mechanisms across models with approximately 70M, 160M, and 500M parameters. Fig.~\ref{fig:scaling} illustrates the predictive performance of these models on the \textsc{WikiText}, \textsc{ARC-E}, and \textsc{SciQ} datasets. Our results indicate that the predictive performance of LMs with a \textit{hybrid} configuration consistently improves with increasing model size. For instance, the accuracy of the \textsl{Non-approximate} method on \textsc{ARC-E} improves from 34.3 to 42.3 when increasing the model size from 70M to 500M. Furthermore, all attention mechanisms incorporating token mixing achieve predictive performance comparable to a same-sized model employing \textsl{standard} attention (indicated by the vertical dotted lines in Fig.~\ref{fig:scaling}). For \textsl{RndEmbQK}, such performance gap on \textsc{WikiText} PPL is even within 1.2 across all sizes. This trend suggests that our observations may generalize to larger models.

To further investigate the immediate generalizability of our findings, we further pretrain a larger model~\citep[Qwen3-1.7b-Base]{yang2025qwen3} from scratch on 45 billion tokens with \textsl{Standard} and our proposed \textsl{RndEmbQK} and \textsl{Non-approximate} variants in both uniform and hybrid configurations. Tbl.~\ref{table:down_stream_performance_xl} presents their performance on NLU and LM tasks. We find both \textsl{RndEmbQK} and \textsl{Non-apx.} under \textit{hybrid} configuration, achieve performance comparable to \textsl{Standard} across all downstream tasks, which is consistent to our observation on models with modest scales. However, different to the model with 500M parameters, \textsl{Non-approximate} under \textit{uniform} configuration successfully converges. This is because Qwen3 incorporates RSMNorm above the queries and key in its attention module. This normalization helps to alleviate the potential for pre-softmax attention activations to explode, but it is less effective than using several standard layers, as it restricts the length of query and key vectors, narrowing the adaptable range for raw pre-softmax activations.


\begin{table*}[!t]
\centering
\small
\resizebox{1.0\linewidth}{!}{
\begin{NiceTabular}{clccccccccrrr}
\toprule
& & \textbf{ARC-E} & \textbf{BoolQ} & \textbf{COPA} & \textbf{PiQA} & \textbf{SciQ} & \textbf{RTE} & \textbf{HellaSwag} & \text{Avg.} & \textbf{Wiki} & \multicolumn{2}{c}{\textbf{LAMBADA}}\\ 
& & acc$\uparrow$ & acc$\uparrow$ & acc$\uparrow$ & acc$\uparrow$ & acc$\uparrow$ & acc$\uparrow$ & acc$\uparrow$ & acc$\uparrow$ & ppl$\downarrow$ & ppl$\downarrow$ & acc$\uparrow$ \\ 
 \cmidrule(lr){3-10}\cmidrule(lr){11-13}

&\textbf{Rnd. Guess} & 25.0\stddev{0.0} & 50.0\stddev{0.0} & 50.0\stddev{0.0} & 50.0\stddev{0.0} & 25.0\stddev{0.0} & 50.0\stddev{0.0} & 25.0\stddev{0.0} & 39.9 & 3E+5 & 3E+6 & 0.0\stddev{0.0} \\ 
&\textbf{Majority} & 25.7\stddev{0.0} & 62.2\stddev{0.0} & 56.0\stddev{0.0} & 50.5\stddev{0.0} & 25.0\stddev{0.0} & 52.7\stddev{0.0} & 25.0\stddev{0.0} & 39.9 & - & - & -\\ \hdashline

&Standard & 44.6\stddev{1.0} & 56.4\stddev{0.9} & 64.0\stddev{4.8} & 64.0\stddev{1.1} & 67.3\stddev{1.5} & 52.7\stddev{3.0} & 30.5\stddev{0.5} & 54.2 & 27.6 & 60.0 & 28.9\stddev{0.6}\\ \midrule

\rowcolor{TOKMIX} \cellcolor{nocolor}{\multirow{2}{*}{\rotatebox[origin=c]{90}{UNI.}}}
\rowcolor{MATHFORM} \cellcolor{nocolor}&Non-apx. & 41.0\stddev{1.0} & 61.9\stddev{0.9} & 58.0\stddev{5.0} & 59.5\stddev{1.2} & 56.9\stddev{1.6} & 52.4\stddev{3.0} & 27.8\stddev{0.5} & 51.1 & 67.3 & 619.6 & 8.1\stddev{0.4}\\
\rowcolor{SEQDEP} \cellcolor{nocolor}&RndEmbQK & 44.4\stddev{1.0} & 50.2\stddev{0.9} & 60.0\stddev{4.9} & 62.4\stddev{1.1} & 56.3\stddev{1.6} & 54.9\stddev{3.0} & 28.6\stddev{0.5} & 51.0 & 54.9 & 1872.8 & 3.8\stddev{0.3}\\
\midrule

\rowcolor{TOKMIX} \cellcolor{nocolor}\multirow{2}{*}{\rotatebox[origin=c]{90}{HYB.}}
\rowcolor{MATHFORM} \cellcolor{nocolor}&Non-apx. & \textbf{45.0}\stddev{1.0} & 58.1\stddev{0.9} & 65.0\stddev{4.8} & 63.4\stddev{1.1} & \textbf{66.2}\stddev{1.5} & 53.1\stddev{3.0} & \textbf{30.2}\stddev{0.5} & 54.4 & 29.9 & 77.4 & \textbf{26.8}\stddev{0.6}\\
\rowcolor{SEQDEP} \cellcolor{nocolor}&RndEmbQK & 45.4\stddev{1.0} & 57.0\stddev{0.9} & 67.0\stddev{4.7} & 64.5\stddev{1.1} & 65.5\stddev{1.5} & 55.2\stddev{3.0} & 30.4\stddev{0.5} & 55.0 & 28.0 & 61.6 & 29.8\stddev{0.6}\\

\bottomrule

\end{NiceTabular}
}
\caption{Performance of \textit{uniform}, \textit{hybrid} and \textsl{standard} models (1.7B). Blue (Non-apx.) and green (RndEmbQK) denote variants that relax \texttt{Mathematical Form}, and \texttt{Sequence-Dependency}, respectively.}
\label{table:down_stream_performance_xl}

\end{table*}

\section{Grouped-query Attention Ablation}
To confirm the generality of our main investigations, we also trained 500M parameter versions of the $\textsl{Standard}$, $\textsl{Non-approximate}$, and $\textsl{RndEmbQK}$ models using the grouped-query attention configuration. These models are trained on the same 15 billion tokens, with precisely matched parameter counts. We observe that the results on downstream tasks remain consistent across both the multi-head attention and grouped-query attention configurations. Their performance on both NLU and LM tasks is detailed in Tbl.~\ref{table:down_stream_performance_gqa}.

\begin{table*}[!t]
\centering
\small
\resizebox{1.0\linewidth}{!}{
\begin{NiceTabular}{clccccccccrrr}
\toprule
& & \textbf{ARC-E} & \textbf{BoolQ} & \textbf{COPA} & \textbf{PiQA} & \textbf{SciQ} & \textbf{RTE} & \textbf{HellaSwag} & \text{Avg.} & \textbf{Wiki} & \multicolumn{2}{c}{\textbf{LAMBADA}}\\ 
& & acc$\uparrow$ & acc$\uparrow$ & acc$\uparrow$ & acc$\uparrow$ & acc$\uparrow$ & acc$\uparrow$ & acc$\uparrow$ & acc$\uparrow$ & ppl$\downarrow$ & ppl$\downarrow$ & acc$\uparrow$ \\ 
 \cmidrule(lr){3-10}\cmidrule(lr){11-13}

&\textbf{Rnd. Guess} & 25.0\stddev{0.0} & 50.0\stddev{0.0} & 50.0\stddev{0.0} & 50.0\stddev{0.0} & 25.0\stddev{0.0} & 50.0\stddev{0.0} & 25.0\stddev{0.0} & 39.9 & 3E+5 & 3E+6 & 0.0\stddev{0.0} \\ 
&\textbf{Majority} & 25.7\stddev{0.0} & 62.2\stddev{0.0} & 56.0\stddev{0.0} & 50.5\stddev{0.0} & 25.0\stddev{0.0} & 52.7\stddev{0.0} & 25.0\stddev{0.0} & 39.9 & - & - & -\\ \hdashline

&Standard & 39.4\stddev{1.0} & 49.6\stddev{0.9} & 60.0\stddev{5.0} & 62.2\stddev{1.1} & 59.3\stddev{1.6} & 51.6\stddev{3.0} & 28.1\stddev{0.5} & 50.0 & 38.6 & 154.0 & 22.9\stddev{0.6}\\ \midrule

\rowcolor{TOKMIX} \cellcolor{nocolor}{\multirow{2}{*}{\rotatebox[origin=c]{90}{UNI.}}}
\rowcolor{MATHFORM} \cellcolor{nocolor}&Non-apx. & 26.8\stddev{0.9} & 37.8\stddev{0.8} & 52.0\stddev{5.0} & 52.0\stddev{1.2} & 20.3\stddev{1.3} & 52.7\stddev{3.0} & 25.9\stddev{0.4} & 38.2 & 5466.8 & 2E+6 & 0.0\stddev{0.0}\\
\rowcolor{SEQDEP} \cellcolor{nocolor}&RndEmbQK & 37.9\stddev{1.0} & 53.2\stddev{0.9} & 56.0\stddev{5.0} & 58.3\stddev{1.2} & 46.7\stddev{1.6} & 52.7\stddev{3.0} & 27.2\stddev{0.4} & 47.4 & 84.6 & 6462.7 & 12.4\stddev{0.2}\\

\rowcolor{TOKMIX} \cellcolor{nocolor}\multirow{2}{*}{\rotatebox[origin=c]{90}{HYB.}}
\rowcolor{MATHFORM} \cellcolor{nocolor}&Non-apx. & 40.7\stddev{1.0} & 44.6\stddev{0.9} & 67.0\stddev{5.0} & 61.3\stddev{1.1} & 61.5\stddev{1.5} & 52.4\stddev{3.0} & 28.3\stddev{0.5} & 50.8 & 38.1 & 133.1 & 23.4\stddev{0.6}\\
\rowcolor{SEQDEP} \cellcolor{nocolor}&RndEmbQK & 40.1\stddev{1.0} & 45.8\stddev{0.9} & 63.0\stddev{4.9} & 61.3\stddev{1.1} & 61.8\stddev{1.5} & 52.7\stddev{3.0} & 28.3\stddev{0.5} & 50.4 & 39.3 & 138.6 & 23.6\stddev{0.6}\\

\bottomrule

\end{NiceTabular}
}
\caption{Performance of \textit{uniform}, \textit{hybrid} and \textsl{standard} models (500m) using grouped-query attention. Blue (Non-apx.) and green (RndEmbQK) denote variants that relax \texttt{Mathematical Form}, and \texttt{Sequence-Dependency}, respectively.}
\label{table:down_stream_performance_gqa}

\end{table*}


\section{Robustness to Context Length}\label{appendix:length}

Tbl.~\ref{table:length_wiki} illustrates the perplexity scores of the \textsc{uniform}, \textsc{hybrid} and \textsl{tandard} models on \textsc{WikiText} dataset. These models were evaluated across various contextual lengths (128, 256, 512, 1024, and 2048 tokens), all while being trained on a maximum sequence length of 2048 tokens. The results clearly show that models incorporating token mixing achieve lower perplexity scores with longer contexts. This indicates their ability to capture more contextual information for predicting the next token. Furthermore, under the \textit{hybrid} configuration, the perplexity scores for the \textsl{RndEmbQK}, \textsl{FixedSeqQK}, \textsl{StaticEmbQK}, \textsl{Approximate} and \textsl{Non-approximate} attention mechanisms consistently match those of the \textsl{standard} model on \textsc{WikiText}, regardless of contextual length.

\begin{table}[!t]
\centering
\small
\resizebox{1.0\linewidth}{!}{
\begin{tabular}{clrrrrr}
\toprule
\textbf{PPL} & \textbf{length} & \textbf{128} & \textbf{256} & \textbf{512} & \textbf{1024} & \textbf{2048}\\ 
 \cmidrule(lr){3-7}

&Standard & 69.9 & 56.2 & 47.8 & 42.0 & 38.1\\ \midrule

\multirow{6}{*}{\rotatebox[origin=c]{90}{UNIFORM}}
&MLP & 993.5 & 993.5 & 993.5 & 993.5 & 993.5\\
&Approx. & 81.7 & 66.4 & 57.2 & 51.2 & 47.9\\
&Non-apx. & 10023.7 & 9476.8 & 9173.5 & 9064.8 & 9025.9\\
&RndEmbQK & 107.1 & 95.7 & 89.6 & 86.3 & 84.8\\
&FixedSeqQK & 100.5 & 89.6 & 83.6 & 80.6 & 79.1\\
&StaticEmbQK & 104.8 & 92.2 & 85.5 & 81.7 & 79.9\\
\midrule

\multirow{6}{*}{\rotatebox[origin=c]{90}{HYBRID}}
&MLP & 81.3 & 66.4 & 57.0 & 50.3 & 45.8\\
&Approx. & 71.8 & 57.8 & 49.3 & 43.4 & 39.4\\
&Non-apx. & 69.0 & 56.0 & 48.0 & 42.5 & 39.4\\
&RndEmbQK & 72.4 & 58.1 & 49.4 & 43.4 & 39.3\\
&FixedSeqQK & 69.0 & 56.0 & 48.0 & 42.3 & 38.5\\
&StaticEmbQK & 70.1 & 56.6	& 48.4 & 42.6 & 38.7\\

\bottomrule

\end{tabular}
}
\caption{Perplexities of \textsl{uniform}, \textsl{hybrid} and \textsl{standard} models (500M) on \textsc{WikiText} across different context lengths.}
\label{table:length_wiki}

\end{table}

\section{Characteristics of Different Simpler Attentions}\label{appendix:characteristics}
Unlike previous work that primarily focused on reducing computational time complexity to sub-quadratic with respect to contextual sequence length, we define ``simpler attention'' more broadly. This encompasses mechanisms that reduce time complexity concerning any factor: inference batch size, sequence length, or hidden dimension. Below, we systematically summarize the characteristics of the different simpler attention mechanisms we investigated.

\paragraph{RndEmbQK and FixedSeqQK.} These mechanisms create global static attention graphs during inference. This approach reduces the computational time complexity and cache size within attention while enabling batched decoding (see Appx.~\ref{appendix:complexity} and~\ref{appendix:cache_size}).

\paragraph{StaticEmbQK.} Inspired by cross-layer attention sharing~\citep{rajabzadeh2024echoatt,xiao2019sharing}, this mechanism primarily captures semantic similarities between input tokens without contextualization. It establishes an upper bound for broadcasting attention matrices from initial layers to all subsequent layers by aligning its parameter count with standard attention. While \textsl{StaticEmbQK} attention does not explicitly reduce computational time complexity, it allows for system optimization by computing attention scores asynchronously. This enables scores to be prefetched before sequentially retrieving output hidden states from each layer.

\paragraph{Approximate and Non-approximate.} These attention mechanisms result in time complexities linear to sequence length. Their recurrent forms are detailed in Appx.~\ref{appendix:recurrent_form}. \textsl{Non-approximate} can further reduce the activation memory, cache size, and floating-point operations per iteration (FLOPs/it) required for large LMs during the decode stage, offering advantages over \textsl{Approximate}. The details for these reductions are provided in Appendices~\ref{appendix:activation},~\ref{appendix:cache_size}, and~\ref{appendix:flop_per_it}, respectively.

\section{Time Complexities in Attention Computation}\label{appendix:complexity}

Tbl.~\ref{table:time_complexities} details the computational time complexity for a single forward pass, explicitly excluding any caching mechanisms. For \textsl{RndEmbQK} and \textsl{FixedSeqQK} attention, which employ global attention scores, the floating-operations could be further reduced to through pre-computation and subsequent caching of these scores (see Appx.~\ref{appendix:flop_per_it}). This optimization would free up computational resources, enabling further software-level enhancements such as coordinating CPUs and GPUs to pre-fetch the pre-calculated attention scores. While \textsl{StaticEmbQK} does not inherently offer a lower computational time complexity, it provides an upper bound for pre-computing attention scores on static embeddings by aligning the number of parameters. If attention scores on static embeddings are pre-computed, the computational time complexity would be reduced by $O\left((l-1)\cdot(BL^2d+BLd^2)\right)$ in total, where $l$ represents the total number of Transformer layers. Furthermore, an attention mechanism that supports pre-computation offers the potential to proactively evict values, which could lead to further reductions in computation, particularly if the attention matrices exhibit sparsity.

\renewcommand*{\arraystretch}{1.0}
\begin{table}[!t]
\begin{center}
\tiny
\resizebox{\linewidth}{!}{
\begin{tabular}{lr}
\toprule
\textbf{Attention} & \textbf{Complexity $\mathcal{O}(.)$}\\
\midrule
Standard & $BL^2d+BLd^2$\\ \hdashline
MLP & $BLd^2$\\
Approx. & $BLd^2$\\
Non-apx. & $BLd^2$\\
RndEmbQK & $BL^2d+BLd^2$\\
FixedSeqQK & $BL^2d+BLd^2$\\
StaticEmbQK & $BL^2d+BLd^2$\\
\bottomrule
\end{tabular}}
\caption{Details of time complexities for each attention across all attention variants, where $h$ denotes the number of attention heads, $B$ denotes the batch size, $L$ denotes the input sequence length, $d$ denotes the hidden dimension. We assume $d=h\times{d_h}$, where $h$ is the number of attention heads and $d_h$ is the dimension of each attention head. We also ignore those low-order terms for element-wise activations and scaling factors with a $\mathcal{O}(BLd)$ complexity.}
\label{table:time_complexities}
\end{center}
\end{table}

\section{Floating-point Operations per Token}\label{appendix:flop_per_it}

Tbl.~\ref{table:float_per_iteration} details the floating-point operations per iteration (FLOP/it) for inference with the cache enabled. We focus solely on General Matrix Multiplications (GEMMs)~\citep[GEMMs]{narayanan2021efficient}, as they are the dominant contributors to the total floating-point operations.

\textsl{Non-approximate} achieves a low FLOP/it, equivalent to that of the simplest \textsl{MLP} model, because it leverages vectors instead of the matrices employed by the Approximate method for state tracking. This structural difference significantly reduces the number of GEMMs required.

Furthermore, if \textsl{RndEmbQK} and \textsl{FixedSeqQK} are allowed to use pre-computed global attention scores, their FLOP/it can be further reduced. During the prefill stage, the operations drop to $2L^2d+2BLd^2$ and $2BLd+2d^2$ during prefill and decode stage respectively.

\renewcommand*{\arraystretch}{1.0}
\begin{table}[!t]
\begin{center}
\tiny
\resizebox{\linewidth}{!}{
\begin{tabular}{lrr}
\toprule
\textbf{Attention} & \textbf{Prefill} & \textbf{Decode}\\
\midrule
Standard & $4BL^2d+6BLd^2$ & $6Bd^2 + 4BLd$\\ \hdashline
MLP & $6BLd^2$ & $6Bd^2$\\
Approx. & $14BLd^2$ & $10Bd^2$\\
Non-apx. & $6BLd^2$ & $6Bd^2$\\
RndEmbQK & $2L^2d+2BL^2d+6BLd^2$ & $2Ld+2BLd+6Bd^2$\\
FixedSeqQK & $2L^2d+2BL^2d+6BLd^2$ & $2Ld+2BLd+6Bd^2$\\
StaticEmbQK & $4BL^2d+6BLd^2$ & $6Bd^2 + 4BLd$\\
\bottomrule
\end{tabular}}
\caption{Details of floating-point operations per iteration for each attention across all attention variants, where $h$ denotes the number of attention heads, $B$ denotes the batch size, $L$ denotes the input sequence length, $d$ denotes the hidden dimension. We assume $d=h\times{d_h}$, where $h$ is the number of attention heads and $d_h$ is the dimension of each attention head.}
\label{table:float_per_iteration}
\end{center}
\end{table}

\section{Activation Memory Required for Attention Computation}\label{appendix:activation}

We detail the activation memory required for half-precision training in Tbl.~\ref{table:activation_memory}. Unlike the full recomputation method mentioned in \citet{smith2022using}, our approach incorporates sequence parallelism following \citet{korthikanti2023reducing}. We find that \textsl{RndEmbQk} and \textsl{FixedSqeQK} are effective at reducing activation memory, particularly when using a substantially large batch size. Furthermore, both \textsl{Approximate} and \textsl{Non-approximate} enhance memory efficiency for long-context processing. \textsl{Non-approximate} offers a superior reduction in activation memory compared to \textsl{Approximate}, especially for large LMs characterized by a relatively large hidden state dimension.

\renewcommand*{\arraystretch}{2.0}
\begin{table}[!t]
\begin{center}
\tiny
\resizebox{\linewidth}{!}{
\begin{tabular}{lr}
\toprule
\textbf{Attention} & \textbf{Activation memory}\\
\midrule
Standard & $\frac{8BLd + 2BL^2h}{t}$\\ \hdashline
MLP & $\frac{8BLd}{t}$\\
Approx. & $\frac{11BLd}{t} + \frac{3Bd^2}{ht}$\\
Non-apx. & $\frac{8BLd + 4BLh}{t}$\\
RndEmbQK & $\frac{4BLd +8Ld + 2L^2h}{t}$\\
FixedSeqQK & $\frac{4BLd + 8Ld + 2L^2h}{t}$\\
StaticEmbQK & $\frac{8BLd + 2BL^2h}{t}$\\
\bottomrule
\end{tabular}}
\caption{Details of activation memory for each attention across all attention variants, where $h$ denotes the number of attention heads, $B$ denotes the batch size, $L$ denotes the input sequence length, $d$ denotes the hidden dimension, $t$ denotes the tensor parallel size. We assume $d=h\times{d_h}$, where $h$ is the number of attention heads and $d_h$ is the dimension of each attention head. We ignore the attention dropout here.}
\label{table:activation_memory}
\end{center}
\end{table}

\section{Cache Size Required for Inference}\label{appendix:cache_size}
Tbl.~\ref{table:infer_cache_size} presents the cache size required for half-precision inference. Both the \textsl{Approximate} and \textsl{Non-approximate} variants allow the cache size to be independent of the context sequence length. Meanwhile, \textsl{RndEmbQk} and \textsl{FixedSeqQK} can reduce the cache size by nearly half by sharing the same set of keys within the same batch, provided the batch size is sufficiently large. It is also important to note that \textsl{RndEmbQK} and \textsl{FixedSeqQk} enable a cache size further optimized to $(2L+\delta)\delta$. This can be achieved by using a dynamic cache and prefetching the attention scores for the next $\delta$ steps into a buffer, given that the attention matrices are independent of the inputs.

\renewcommand*{\arraystretch}{1.0}
\begin{table}[!t]
\begin{center}
\tiny
\resizebox{\linewidth}{!}{
\begin{tabular}{lr}
\toprule
\textbf{Attention} & \textbf{Cache Size for Inference}\\
\midrule
Standard & $4BLd$\\ \hdashline
MLP & $0$\\
Approx. & $6Bd + 4Bd^2/h$\\
Non-apx. & $2Bd + 4Bh$\\
RndEmbQK & $2(B+1)Ld$\\
FixedSeqQK & $2(B+1)Ld$\\
StaticEmbQK & $4BLd$\\
\bottomrule
\end{tabular}}
\caption{Details of cache size (in bytes) per layer across all attention variants required during inference, where $h$ denotes the number of attention heads, $B$ denotes the batch size, $L$ denotes the context length, $d$ denotes the hidden dimension. We assume $d=h\times{d_h}$, where $h$ is the number of attention heads and $d_h$ is the dimension of each attention head.}
\label{table:infer_cache_size}
\end{center}
\end{table}

\section{Recurrent Form of Linear Attentions}\label{appendix:recurrent_form}

The recurrent form of the \textit{Approximate} attention computation, derived from Eq.~\ref{eq:approx_attn}, is presented in Eq.~\ref{eq:recurrent_approx}. Similarly, Eq.~\ref{eq:recurrent_nonapprox} shows the recurrent form of the \textit{Non-approximate} attention computation, originating from Eq.~\ref{eq:non_approx_attn}. As detailed in Tbl.~\ref{table:time_complexities}, the \textit{Approximate} attention mechanism necessitates the computation of recursions for both first-order and second-order terms in the Taylor expansion, resulting in a higher time complexity compared to the \textit{Non-approximate} approach. A key characteristic of $\mathbf{O}_i$ in Eq.~\ref{eq:recurrent_nonapprox} is that its denominator strictly increases with the index $i$. Notably, as $i$ grows along the sequence, the attention score for the $i^\text{th}$ token, given by $\frac{\mathrm{e}^{q_ik_i^{\top}}v_i}{\sum^{i-1}_{j=1}\mathrm{e}^{q_jk_j^{\top}}+\mathrm{e}^{q_ik_i^{\top}}}$, becomes progressively more challenging to increase.

{
\begin{align}
    \mathbf{o}_i&=\mathbf{o}_{0i}+\mathbf{o}_{1i}+\mathbf{o}_{2i}\\
    \mathbf{o}_{0i}&=\frac{\sum^{i-1}_{j=1}v_j+v_i}{i}\\
    \mathbf{o}_{1i}&=\frac{q_i\left(\sum^{i-1}_{j=1}k_j^\top{v_j}+k_i^\top{v_i}\right)}{q_i\left(\sum^{i-1}_{j=1}k_j^\top+k_i^\top\right)}\\
    \mathbf{o}_{2i}&=\frac{\frac{q_i^2}{\sqrt{2}}\left(\sum^{i-1}_{j=1}(\frac{k_j^2}{\sqrt{2}})^{\top}{v_j}+(\frac{k_i^2}{\sqrt{2}})^{\top}{v_i}\right)}{\frac{q_i^2}{\sqrt{2}}\left(\sum^{i-1}_{j=1}(\frac{k_j^2}{\sqrt{2}})^{\top}+(\frac{k_i^2}{\sqrt{2}})^{\top}\right)}
    \label{eq:recurrent_approx}
\end{align}
}

{
\begin{align}
    \mathbf{o}_i&=\frac{\sum^{i-1}_{j=1}\mathrm{e}^{q_jk_j^{\top}}v_j+\mathrm{e}^{q_ik_i^{\top}}v_i}{\sum^{i-1}_{j=1}\mathrm{e}^{q_jk_j^{\top}}+\mathrm{e}^{q_ik_i^{\top}}}\label{eq:recurrent_nonapprox}
\end{align}
}

\section{Hyperparameters}\label{appendix:hyperparameters}
The hyperparameters used in pre-training are listed in Tbl.~\ref{table:hyperparams_pretraining}.

\renewcommand*{\arraystretch}{1.0}
\begin{table}[!t]
\begin{center}
\small
\resizebox{\linewidth}{!}{
\begin{tabular}{ll}
\toprule
\multicolumn{2}{c}{Hyperparameters in Pretraining}\\ \midrule
Maximum train steps & 120000\\
Batch size (in total) & 256 instances\\
Adam $epsilon$ & 1e-8\\
Adam $\beta_1$ & 0.9\\
Adam $\beta_2$ & 0.9999\\
Sequence length & 2048\\
Peak learning rate & 4e-4 (3e-4 for Qwen3-1.7B)\\
Learning rate schedule & CosineLRScheduler\\
Number of cycles in scheduler & 0.5\\
Warmup steps & 2000 (1B tokens)\\
Weight decay & 0.1\\
Max gradient norm clip value & 1.0\\
\bottomrule
\end{tabular}}
\caption{Details of hyperparameters used in pre-training.}
\label{table:hyperparams_pretraining}
\end{center}
\end{table}

\section{Training Loss across all Attention Mechanisms}

\begin{figure}[!t]
\centering
\begin{subfigure}{\linewidth}
\includegraphics[width=\linewidth]{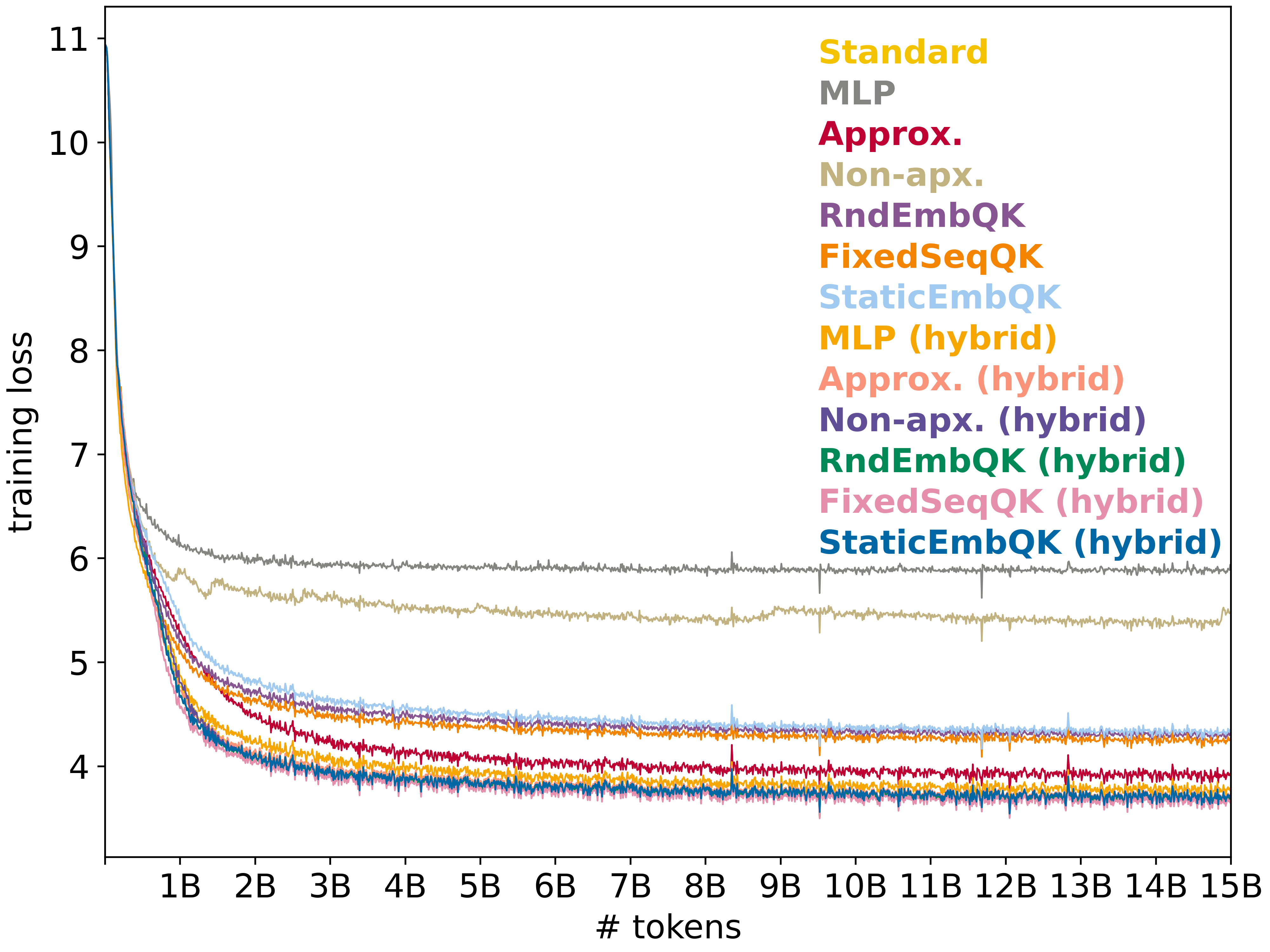}
\caption{Training loss across models with 70M parameters}
\label{fig:loss_70m}
\end{subfigure}

\bigskip

\begin{subfigure}{\linewidth}
\includegraphics[width=\linewidth]{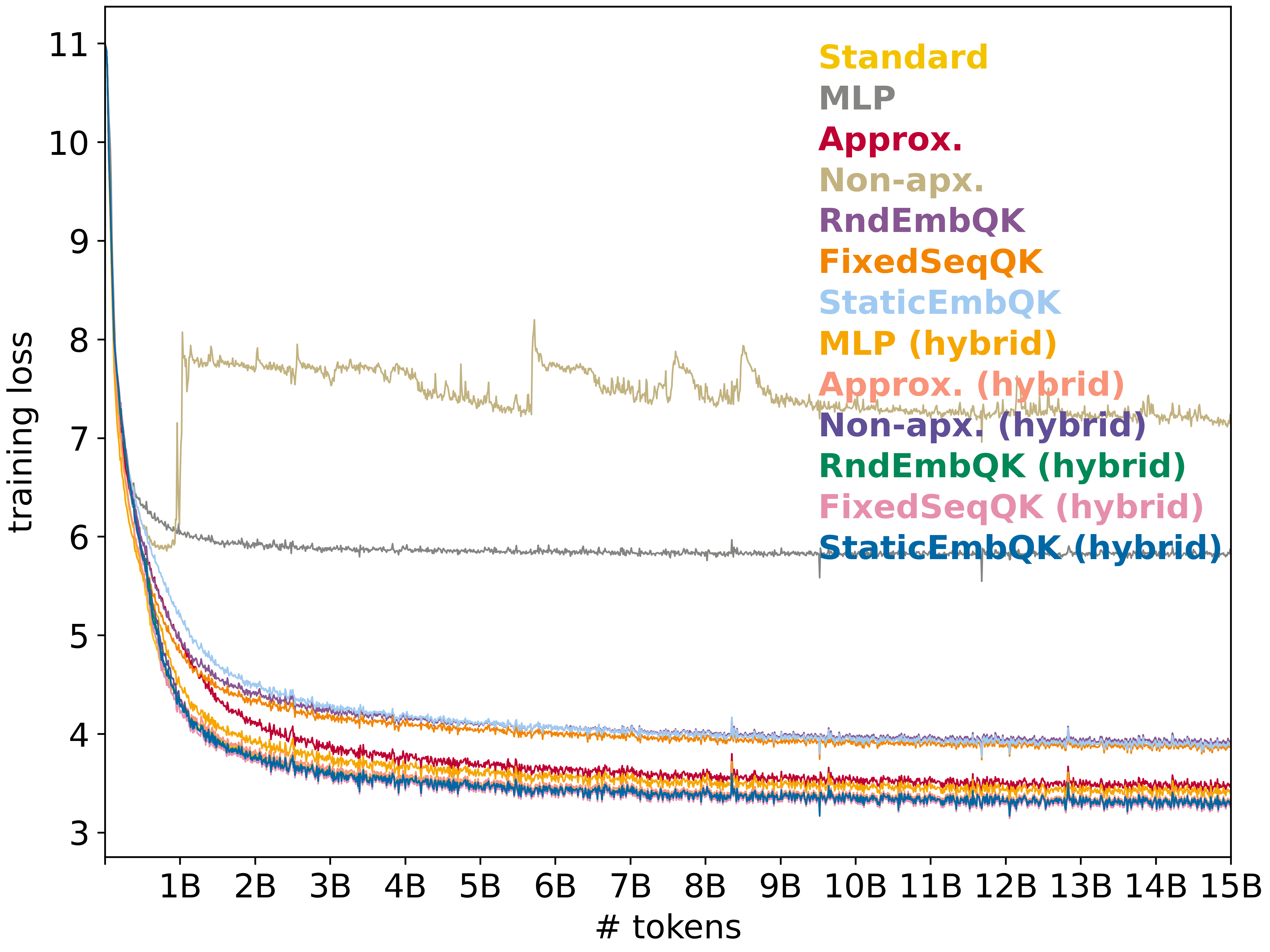}
\caption{Training loss across models with 160M parameters}
\label{fig:loss_160m}
\end{subfigure}

\bigskip

\begin{subfigure}{\linewidth}
\includegraphics[width=\linewidth]{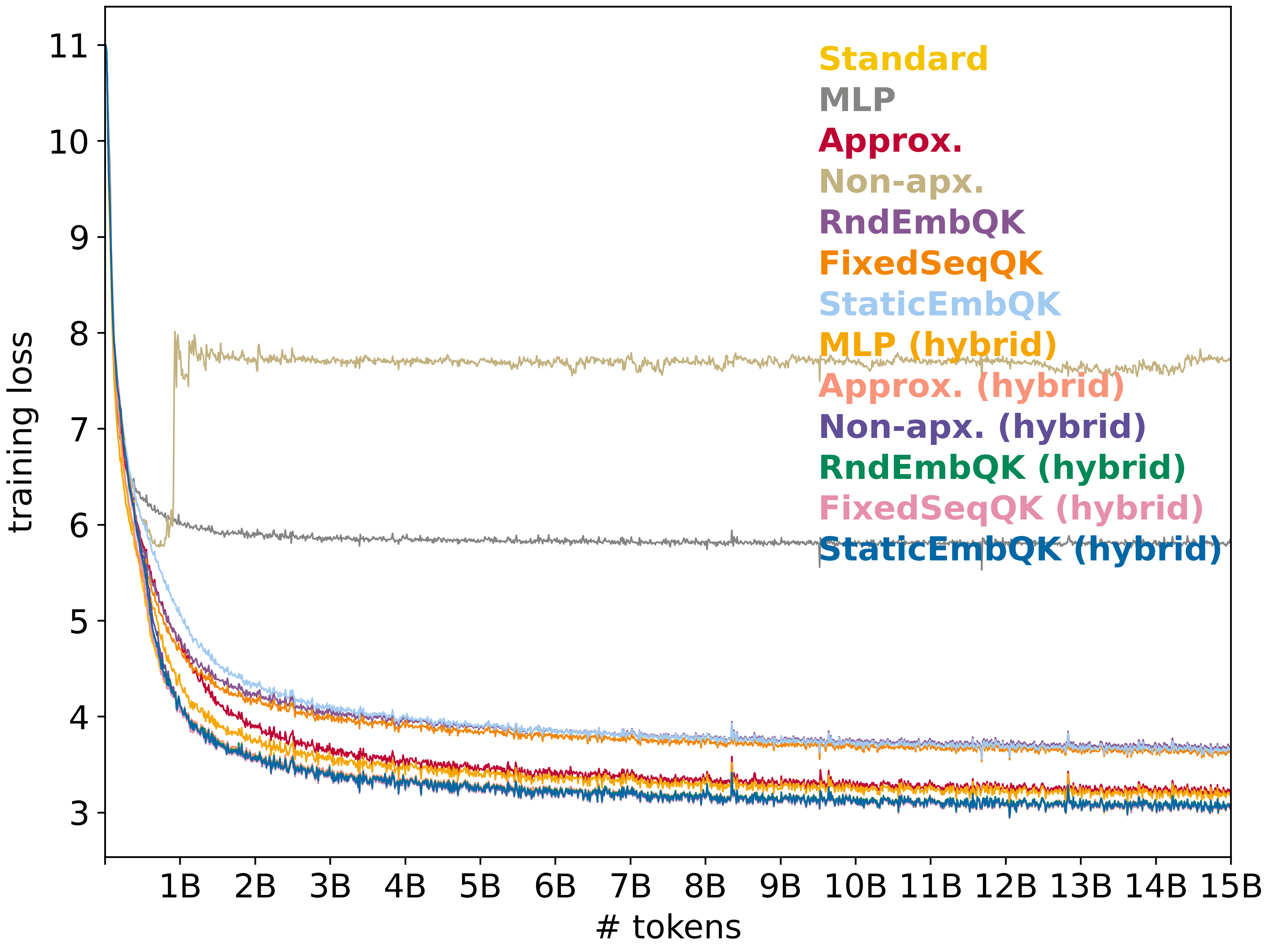}
\caption{Training loss across models with 500M parameters}
\label{fig:loss_500m}

\end{subfigure}

\caption{Training loss across all model variants with three different sizes.}\label{fig:loss}
\end{figure}

Fig.~\ref{fig:loss} presents the loss curves across all model variants and sizes, while training for 15B tokens.

\section{Model Configurations for Different Sizes}\label{appendix:model_config_appendix}

Tbl.~\ref{table:model_config} presents the detailed configurations of models across various sizes (70M, 160M, 500M and 1.7B).

\renewcommand*{\arraystretch}{1.0}
\begin{table}[!t]
\begin{center}
\small
\resizebox{\linewidth}{!}{
\begin{tabular}{lrrrr}
\toprule
Model Size & 70M & 160M & 500M & 1.7B\\ \midrule
Hidden Size & 512 & 768 & 896 & 2048\\
Intermediate Size & 2048 & 3072 & 4864 & 6144\\
Num of Hidden Layers & 6 & 12 & 24 & 28\\
Max Window Layers & 6 & 12 & 24 & 28\\
Num of Attention Heads & 8 & 12 & 14 & 16\\
Num of Key Value Heads & 8 & 12 & 14 & 16\\
\bottomrule
\end{tabular}}
\caption{Details of model configurations for different sizes.}
\label{table:model_config}
\end{center}
\end{table}

\section{Distribution of Raw Logits}

\begin{figure*}[!t]
\centering
\includegraphics[width=\textwidth]{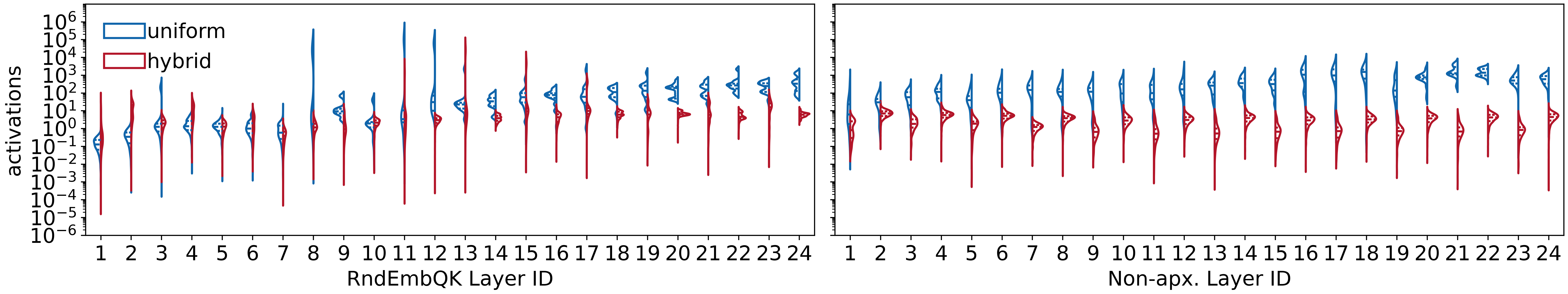}
\caption{Distribution of raw logits in the pre-softmax activations for \textsl{RndEmbQK} (left) and \textsl{Non-approximate} (right) attention mechanisms in both uniform and hybrid configurations.}\label{fig:violin_full}
\end{figure*}

Fig.~\ref{fig:violin_full} (the full version of Fig.~\ref{fig:violin}) exhibits the magnitude of pre-softmax activations within each 24-layer (500M) \textsl{RndEmbQK} and \textsl{Non-approximate} layer in the \textit{hybrid} configuration.

\section{Attention Characteristics from All Layers across Attention Variants}

\begin{figure*}[!t]
\centering
{%
\includegraphics[width=\linewidth]{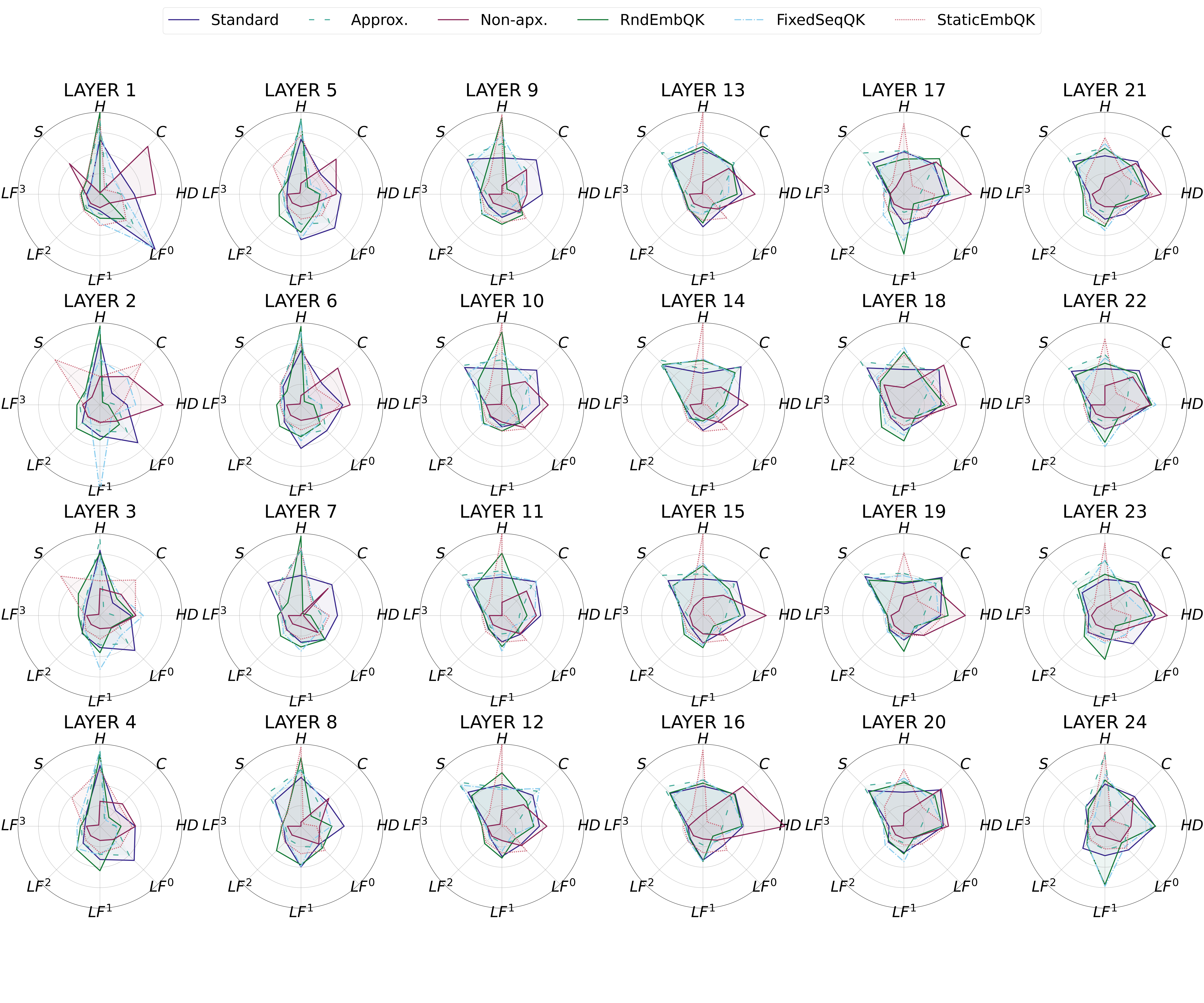}
}%
\caption{Visualization of attention matrix characteristics across different layers for \textsl{Approximate}, \textsl{Non-approximate}, \textsl{RndEmbQK}, \textsl{FixedSeqQK} and \textsl{StaticEmbQK}, and their hybrid variants, compared to \textsl{Standard} ($H$: \textsc{Entropy}, $C$: \textsc{Conc}, $HD$: \textsc{HeadDiv}, $LF$: \textsc{LocFoc$N$}, $S$: \textsc{Sink}).}\label{fig:radar_uniform_all_layers}
\end{figure*}

Fig.~\ref{fig:radar_uniform_all_layers}, the full version of the left subfigure in Fig.~\ref{fig:radar_24}), exhibits attention characterstics from all 24 layers across \textsl{Standard} attention and five attention variants - \textsl{RndEmbQK}, \textsl{FixedSeqQK}, \textsl{StaticEmbQK}, \textsl{Approximate} and \textsl{Non-approximate}.

\section{Attention Characteristics from All Layers across Configurations}

\begin{figure*}[!t]
\centering
{%
\includegraphics[width=\linewidth]{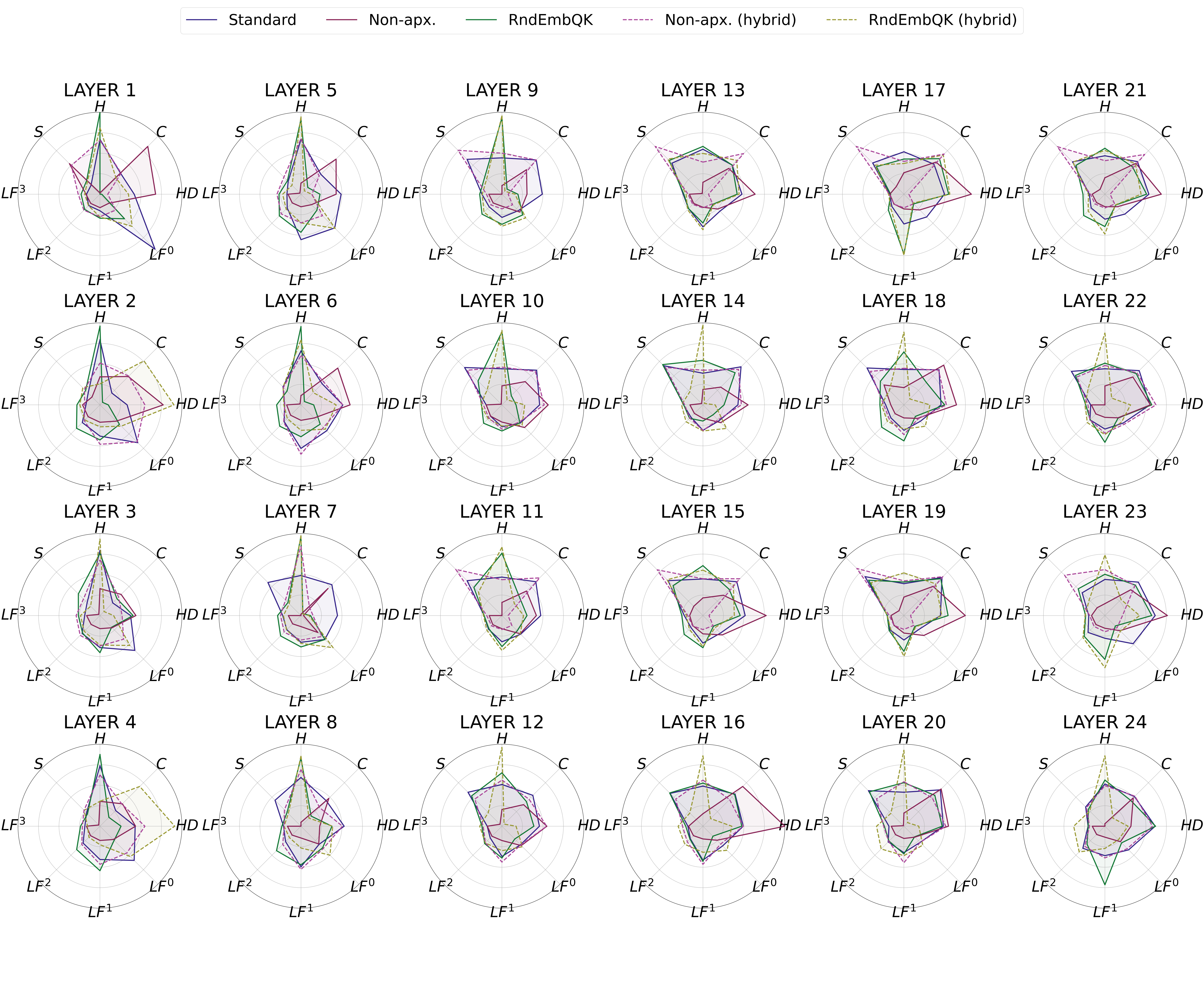}
}%
\caption{Visualization of attention matrix characteristics across different layers for \textsl{Non-approximate} and \textsl{RndEmbQK}, and their hybrid variants, compared to \textsl{Standard} ($H$: \textsc{Entropy}, $C$: \textsc{Conc}, $HD$: \textsc{HeadDiv}, $LF$: \textsc{LocFoc$N$}, $S$: \textsc{Sink}).}\label{fig:radar_hybrid_all_layers}
\end{figure*}

Fig.~\ref{fig:radar_hybrid_all_layers}, the full version of the right subfigure in Fig.~\ref{fig:radar_24}, exhibits attention characterstics from all 24 layers across \textsl{Standard} and two representative attention variants - \textsl{Approximate} and \textsl{Non-approximate} in both \textit{uniform} and \textit{hybrid} configurations.

\section{Model Configurations for Ablation Study}\label{appendix:model_config_appendix_abla}

Tbl.~\ref{table:model_config_abla} details nine distinct \textit{hybrid} architectures, as discussed in \S~\ref{sec:different_hybrid_config}, for 24-layer model variants with approximately 500 million parameters.

\renewcommand*{\arraystretch}{1.0}
\begin{table}[!t]
\begin{center}
\small
\resizebox{\linewidth}{!}{
\begin{tabular}{lcc}
\toprule
Config & Standard Layer IDs \\ \midrule
even (50\%) & \{2,4,6,8,10,12,14,16,18,20,22,24\}\\
odd & \{1,3,5,7,9,11,13,15,17,19,21,23\}\\
top & \{1,2,3,4,5,6,7,8,9,10,11,12\}\\
middle & \{1,2,3,4,5,6,19,20,21,22,23,24\}\\
bottom & \{13,14,15,16,17,18,19,20,21,22,23,24\}\\
25\% & \{4,8,12,16,20,24\}\\
first & \{1\}\\
last & \{24\}\\
bilteral & \{1,24\}\\

\bottomrule
\end{tabular}}
\caption{Details of model configurations for ablation study.}
\label{table:model_config_abla}
\end{center}
\end{table}

\end{document}